\documentclass{article}

\usepackage{microtype}
\usepackage{graphicx}
\usepackage{subcaption}
\usepackage{booktabs} 

\usepackage{hyperref}

\usepackage[preprint]{icml2026}

\usepackage{amsmath}
\usepackage{amssymb}
\usepackage{mathtools}
\usepackage{amsthm}

\usepackage{listings}      
\usepackage{graphicx}      
\usepackage{wrapfig}       
\usepackage{hyperref}      
\usepackage{url}           
\usepackage{booktabs}      
\usepackage{amsfonts}      
\usepackage{nicefrac}     
\usepackage{microtype}    
\usepackage{textcomp}      
\usepackage{rotating}     
\usepackage{multirow}    
\usepackage{array}        
\usepackage{makecell}    
\usepackage[table]{xcolor} 
\usepackage{placeins}      
\usepackage{subcaption}  
\usepackage{marvosym}     
\usepackage{amsmath}     
\usepackage{amssymb}

\makeatletter
\renewcommand{\paragraph}{%
  \@startsection{paragraph}{4}%
  {\z@}{0.5em}{-0.5em}%
  {\normalfont\normalsize\bfseries}%
}

\lstdefinestyle{promptstyle}{
    basicstyle=\footnotesize\ttfamily,
    backgroundcolor=\color{gray!10},
    frame=single,
    breaklines=true,         
    breakatwhitespace=true,  
    breakindent=0pt,         
    captionpos=b,
    keepspaces=true,
    numbers=none,
    numbersep=5pt,
    xleftmargin=0cm,         
    xrightmargin=0cm,       
    showstringspaces=false,
    tabsize=2,
    aboveskip=1em,
    belowskip=1em,
    columns=flexible,        
    linewidth=\linewidth , 
    postbreak=\mbox{\textcolor{gray}{$\hookrightarrow$}\space}
}

\usepackage[capitalize,noabbrev]{cleveref}

\theoremstyle{plain}

\theoremstyle{definition}

\theoremstyle{remark}

\usepackage[textsize=tiny]{todonotes}

\icmltitlerunning{Is Your LLM Really Mastering the Concept? A Multi-Agent Benchmark}

\begin{document}

\twocolumn[
  \icmltitle{Is Your LLM Really Mastering the Concept? A Multi-Agent Benchmark}




\icmlsetsymbol{equal}{*}
\icmlsetsymbol{corr}{\Letter}

\begin{icmlauthorlist}
  \icmlauthor{Shuhang Xu}{bnu}
  \icmlauthor{Weijian Deng}{anu}
  \icmlauthor{Yixuan Zhou}{edu101}
  \icmlauthor{Fangwei Zhong}{bnu,corr}
\end{icmlauthorlist}

\icmlaffiliation{bnu}{Beijing Normal University}
\icmlaffiliation{anu}{Australian National University}
\icmlaffiliation{edu101}{Beijing 101 Education Group}

\icmlcorrespondingauthor{Fangwei Zhong}{fangweizhong@bnu.edu.cn}

  \icmlkeywords{LLM benchmark, concept knowledge}

  \vskip 0.3in
]



\printAffiliationsAndNotice{}  

\begin{abstract}
Concepts serve as fundamental abstractions that support human reasoning and categorization. However, it remains unclear whether large language models truly capture such conceptual structures or primarily rely on surface-level pattern memorization. Existing benchmarks are largely static and fact oriented, which limits their ability to probe fine-grained semantic understanding and makes them vulnerable to data leakage and overfitting. 
To address this limitation, we introduce \textbf{CK-Arena}, a dynamic benchmark for conceptual knowledge evaluation based on a multi agent social deduction game, namely the Undercover game. In this setting, LLM based agents are assigned subtly different concept words and must describe, distinguish, and infer conceptual properties from others’ statements. Model performance is evaluated through both game level outcomes and the semantic quality of generated descriptions.
Furthermore, CK-Arena leverages the interaction process to automatically construct high quality question answering data for fine grained diagnostic analysis. Experimental results show that conceptual understanding varies substantially across models and categories, and is not strictly aligned with overall model capability. The data and code are available at the project homepage: \url{https://ck-arena.site}.

\end{abstract}

\section{Introduction}

Understanding concepts requires recognizing their relationships as well as the similarities and differences that distinguish closely related ones, which is a fundamental aspect of human cognition \citep{wu2012probase,gong2016representing,ji2019microsoft,zhang2021alicg,wang2024-conceptDB2-editing,cao2024survey}. For example, the concept \textit{Primates} unites animals such as \textit{monkeys} and \textit{apes} through features like opposable thumbs, forward-facing eyes, and advanced cognitive abilities, while also involving subtle distinctions such as the presence of tails in most \textit{monkeys} but not in \textit{apes}. Human cognition naturally uses such conceptual structures for reasoning, but it remains unclear to what extent Large Language Models (LLMs) internalize and exploit these abstractions. 

Recent work has highlighted the importance of conceptual knowledge as a core aspect of intelligence. Studies have examined conceptual design generation~\citep{ma2023conceptDB3-generation}, concept editing~\citep{wang2024-conceptDB2-editing}, and abstract concept understanding~\citep{liao2023conceptDB1conceptunderstanding, chen2025conceptDB4-method}, showing growing interest in concept-based reasoning for LLMs. Yet progress remains constrained by the lack of systematic benchmarks.
Traditional benchmarks have advanced LLMs performance~\citep{hendrycks2020mmlu,zellers2019hellaswag,liang2022HELM,mostafazadeh2017lstory}, but most rely on static question–answer formats that test token-level accuracy and factual recall. These evaluations reduce knowledge to isolated items and mainly measure information retrieval, offering little evidence of whether models understand conceptual relationships or can distinguish closely related concepts. For example, a model may identify that \textit{monkeys} and \textit{apes} both belong to \textit{Primates}, but this does not show understanding of the hierarchical relationships or distinctive features between the two groups. Moreover, fixed formats such as multiple-choice questions provide only a partial view of reasoning, and the reliance on static datasets limits scalability, since building and updating them requires extensive annotation.

\begin{figure}[t] 
    \centering
    \includegraphics[width=\columnwidth]{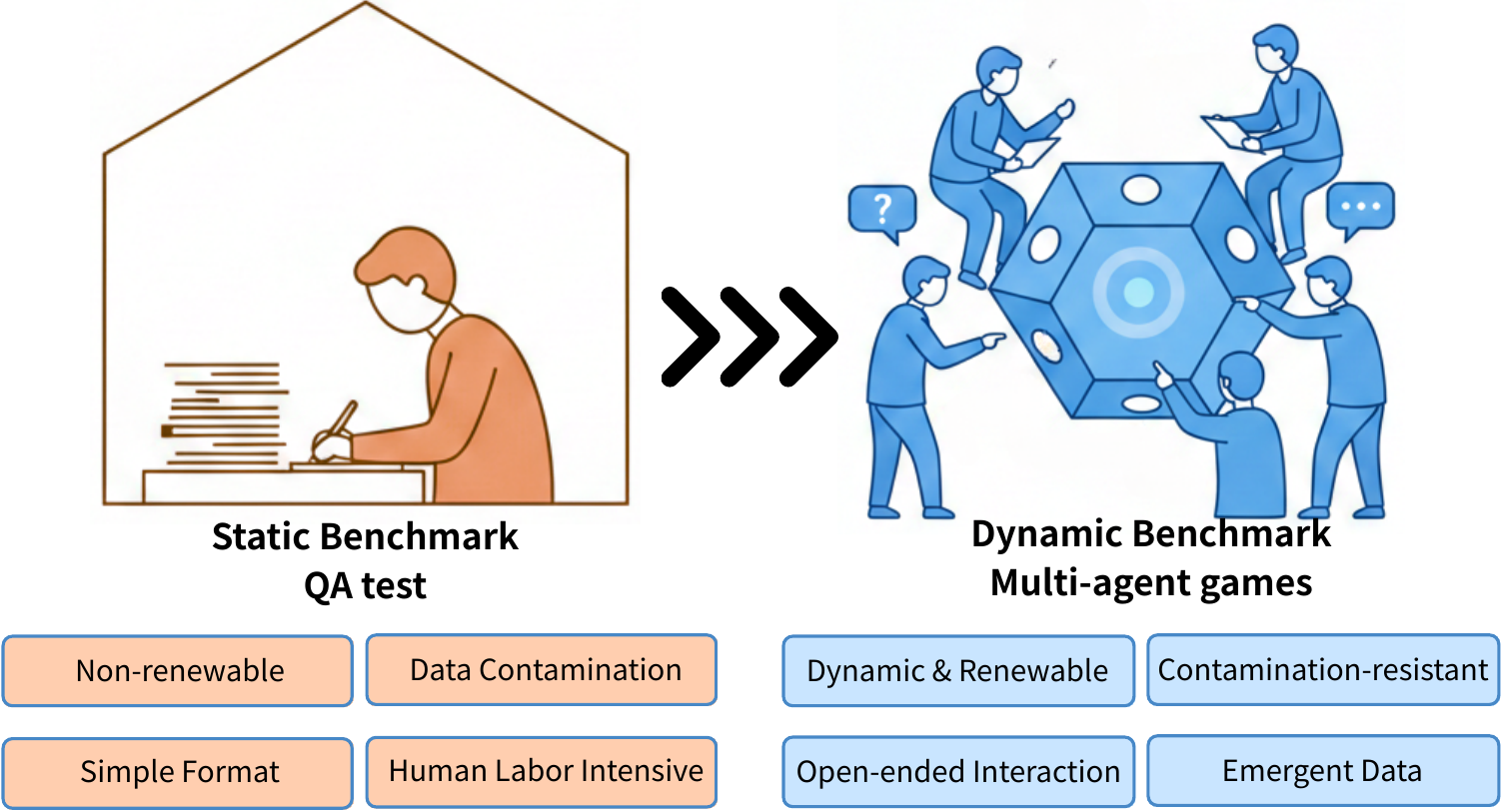}
    \caption{\textbf{Comparison between static and dynamic benchmarks.} Unlike static QA-based evaluations that rely on manually constructed test sets, dynamic benchmarks assess models through interaction with environments and other agents, enabling more realistic evaluation and automatic data generation.
    }
    \label{fig:paradigm}
    \vspace{-1cm}
\end{figure}

Interactive game-based environments have emerged as an alternative \citep{lin2023similar_sandbox,zhou2023sotopia,Wu2023crafter}, 
This is becoming a paradigm shift trend for benchmarks, as shown in Figure \ref{fig:paradigm}.
However, most existing simulations emphasize strategy, providing limited insight into whether models can represent and communicate conceptual knowledge. These gaps call for a systematic and scalable benchmark that directly evaluates conceptual reasoning in realistic interactive settings.

To address this gap, we introduce CK-Arena, an interactive multi-agent benchmark for evaluating the capability of LLMs to represent, differentiate, and communicate conceptual knowledge. We evaluate the LLMs by having them play the \textit{undercover (``Who is the spy?")} game, a multi-agent language game that involves describing a targeting word and identifying each player's role, and by assessing their multi-turn performance as well as their in-game statements. Unlike traditional dataset-based or strategy-focused benchmarks, CK-Arena engages models with concept pairs that share both overlapping and distinctive features, and offers scalable datasets, systematic evaluation protocols, and extensible tools for assessing conceptual understanding.

For evaluation, LLMs serve as referees and are combined with human calibration to ensure reliability. We test a set of recent language models over multiple rounds using a convergent rating system, producing an intuitive leaderboard of their relative performance. Beyond overall ratings, we also analyze results from different perspectives, including in-game success and text generation quality. Experiments show that LLMs’ conceptual understanding varies across categories and does not consistently align with their general capabilities, highlighting the need for targeted evaluation beyond surface-level performance.
To ensure the reliability of game-based evaluation,  we automatically extract game-play data to construct a snapshot QA benchmark comprising fine-grained tasks such as cross-concept inference, fine-grained comparison, and outlier detection. The strong correlation between the QA benchmark performance and game win rates (Spearman $\rho = 0.89$) confirms that dynamic game outcomes genuinely reflect underlying conceptual knowledge.


In summary, our contributions are four-fold: 
(1) we propose CK-Arena, a benchmark for conceptual understanding in interactive multi-agent settings; 
(2) we develop scalable datasets for concept representation, differentiation and connection, along with an automatic pipeline that generates snapshot benchmarks from game-play data for fine-grained diagnosis; 
(3) we conduct a large number of experiments on several models, and obtain knowledge and behavior preferences of specific models through qualitative and quantitative analysis; and (4) we establish a scoring system to integrate fine-grained indicators into a comprehensive score, and launch an overview leaderboard.

\section{Related Works}

\paragraph{Benchmarks for Conceptual Knowledge Reasoning.}
Commonsense reasoning benchmarks play an important role in assessing the capabilities of Large Language Models (LLMs). Widely used benchmarks such as Story Cloze Test \citep{mostafazadeh2017lstory}, Choice of Plausible Alternatives (COPA) \citep{roemmele2011COPA}, and HellaSwag \citep{zellers2019hellaswag} largely rely on static formats like multiple-choice questions or binary judgments. While effective for evaluating factual recall and superficial understanding, these static formats do not fully reflect real-world interactive scenarios. More recent benchmarks, including MMLU \citep{hendrycks2020mmlu}, CMMLU \citep{li2024cmmlu}, BIG-Bench \citep{srivastava2022bigbench}, and HELM \citep{liang2022HELM}, have introduced various tasks such as logical reasoning, cloze tests, and multi-turn Q\&A. Although these efforts represent progress toward more interactive assessments, they still focus predominantly on factual recall and task-specific reasoning, offering limited insight into how well LLMs understand and manipulate conceptual knowledge boundaries in evolving contexts.
In contrast, CK-Arena is designed to explicitly evaluate conceptual mastery by immersing LLMs in interactive, multi-agent gameplay that requires real-time understanding of semantic boundaries.

\begin{figure*}[h!]
    \centering
    \includegraphics[width=1\textwidth]{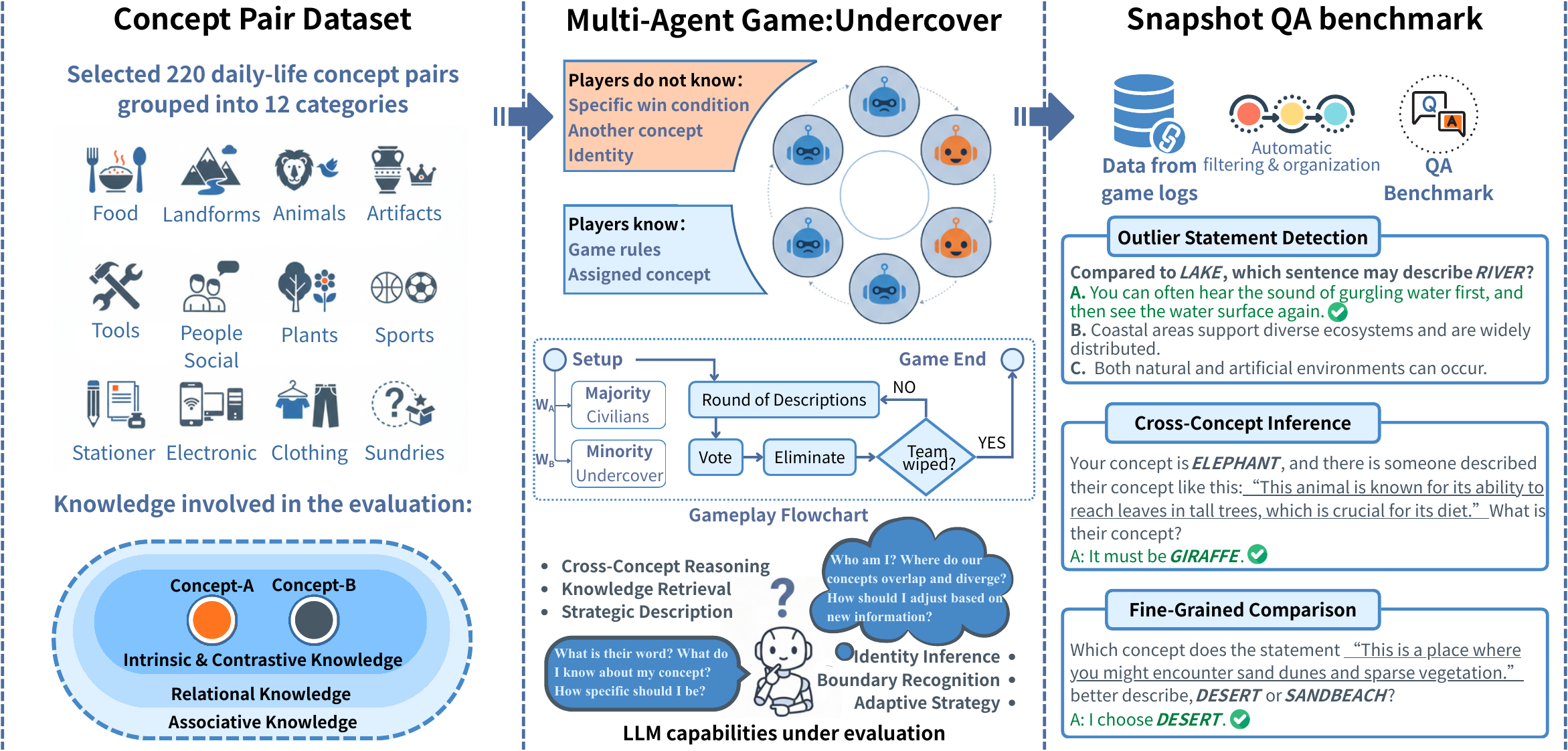} 
    \caption{
    \textbf{CK-Arena aims to evaluate the mastery of conceptual feature knowledge by LLMs.}
    It is built upon the interactive game Undercover, where closely related concept pairs with overlapping and distinguishing attributes are assigned to different LLM agents. Acting as players, models generate descriptions, infer similarities and differences, and make decisions under partial information. Acting as judges, they evaluate responses using sentence-level metrics. This multi-agent design creates a dynamic and scalable setting for assessing conceptual understanding.
    To further support reliable evaluation, we automatically extract gameplay traces to construct a snapshot QA benchmark. This benchmark includes fine-grained tasks such as cross-concept inference, detailed comparison, and outlier detection, enabling systematic analysis of conceptual knowledge across models.
}
    \label{overview}
\end{figure*}

\paragraph{Game-based Evaluation.} 
Multi-agent games offer interactive and dynamic environments that differ from traditional benchmarks built on static datasets. They have been used to measure various skills, including environmental perception and planning in exploratory games \citep{wang2023minecraft, Wu2023crafter}, strategic decision-making in competitive games \citep{NEURIPS2023chessgpt, ma2024starcraft}, team collaboration in cooperative games \citep{agashe2023coop1, mosquera2024coop2}, and social interaction and language comprehension in communication games \citep{light2023avalonbench, qiao2023communication1, wu2024communication2}.
Compared to fixed QA datasets, game-based evaluations better reflect real-world decision making by requiring models to reason, adapt, and respond under uncertainty. 
However, many existing game benchmarks rely on fixed rules or repetitive interaction patterns, limiting their ability to probe deeper reasoning abilities. The Undercover game \citep{xu-etal-2024-magic-undercover} is distinctive in that interchangeable concept pairs naturally induce diverse behaviors within a consistent structure. While prior work has used Undercover to study strategy or coordination \citep{xu-zhong-2025-comet, dong2024undercover, wei2025whoisthespy}, its potential for evaluating conceptual understanding has not been fully explored.
CK-Arena fills this gap by integrating concept-based reasoning within multi-agent interactions, allowing LLMs to explore and articulate conceptual relationships dynamically, mirroring real-world cognitive processing.


\section{CK-Arena: Conceptual Knowledge Arena}

This section introduces the construction of CK-Arena, detailing the choice of the \textit{Undercover} game as the evaluation paradigm, the metrics employed to capture different dimensions of model performance, and the overall workflow for building, running, and analyzing the evaluation.
Together, these components establish CK-Arena as a rigorous and scalable framework for uncovering both the strengths and limitations of LLMs in conceptual knowledge. Figure \ref{overview} shows an overview of CK-Arena.

\subsection{The Undercover Game for Evaluation}

\paragraph{Game Rule.}
CK-Arena is built on the multi-agent language game \textit{Undercover}~\citep{xu-etal-2024-magic-undercover}, which is designed to test players’ reasoning and communication abilities. In the game, players are assigned either as ``civilians",' who are the majority and know a common word, or as ``undercover", who are given a different but related word. Each player is informed of their assigned concept word but remains unaware of their team identity or the concepts held by others. Through rounds of description, players must identify who the undercover agents are while undercover agents try to remain undetected by providing descriptions vague enough to seem plausible without revealing their ignorance of the civilians’ word.
After each round, players participate in a voting process to eliminate the individual they suspect to be an undercover agent. The game concludes under one of two conditions: (1) if all undercover agents are eliminated, the civilians win; (2) if the number of civilians and undercover agents is equal, the undercover agents win.

\paragraph{Why Use Undercover to Evaluate?} 
The core advantage of the \textit{Undercover} game is that changing only the assigned concepts naturally produces new interactions, allowing the same framework to evaluate a wide range of knowledge domains without redesigning tasks. Unlike strategy-oriented games where repeated play mainly reduces randomness, Undercover requires models to reason about conceptual similarity and distinction in each round. 
To illustrate the effectiveness of the \textit{Undercover} game in CK-Arena, consider an example where the concepts \textit{football} and \textit{basketball} are assigned to players, with \textit{basketball} designated as the undercover concept. 
During the speaking phase, the undercover player must analyze the descriptions provided by others about \textit{football}, identify shared attributes, and strategically describe \textit{basketball} in a way that overlaps with common features, such as \textit{``This is a ball-shaped sports equipment''} or \textit{``This sport is played by two teams.''}  
This task requires more than superficial word associations or token co-occurrence. It calls for understanding the similarities and differences between concepts. A model that fails to capture these relationships and relies on shallow generation risks exposing its undercover role and being eliminated.
With its emphasis on conceptual understanding, interactive dynamics, and scalable coverage, CK-Arena provides a rigorous benchmark for evaluating LLMs’ understanding of conceptual knowledge.


\subsection{Large Language Models as Players}

\paragraph{Pipeline.}
In CK-Arena, LLMs participate as players in multi-round games. Each game consists of six agents, including four civilians and two undercover agents. At initialization, civilians receive a shared target concept, while undercover agents are assigned a closely related but distinct concept. Players then take turns generating short descriptions based on their assigned concepts while attempting to infer the roles of others.
Each turn requires the model to perform two key operations. First, it must interpret partial clues from other players and infer the underlying concept based on semantic overlap and divergence. Second, it must retrieve and express relevant features of its own concept in a way that is informative yet strategically appropriate. This process engages both concept-to-concept reasoning and concept-to-feature mapping, allowing CK-Arena to evaluate how well a model understands, distinguishes, and applies conceptual knowledge in an interactive setting.


\paragraph{Prompt Design.}
To ensure effective communication and role-specific behavior, we construct tailored prompts for LLM-based agents in CK-Arena. The prompts include a comprehensive system prompt that provides game rules, input-output format guidelines, specific task instructions, basic strategic guidance, and example descriptions. In addition, each player receives a contextualized user prompt containing information about their assigned concept, historical statements, and analytical insights from previous rounds.
Since CK-Arena is designed to evaluate conceptual mastery, we restrict players’ strategic space with clear action guidelines to avoid confounding effects from uncontrolled reasoning and decision-making.

\subsection{Dynamic Evaluate Protocol}

\paragraph{Data Preparation.}
The selection of concept pairs is critical to the effectiveness of CK-Arena. We construct a dataset of semantically related concept pairs spanning diverse categories. Candidate concepts are drawn from high-frequency vocabulary lists across multiple domains \cite{coca20k}, and are filtered through pilot experiments to ensure two key properties: (1) semantic proximity, meaning the concepts are sufficiently similar to enable challenging discrimination, and (2) descriptive adequacy, meaning each concept can be naturally and clearly expressed during gameplay.

The final dataset contains 529 English concept pairs, including 220 concrete nouns, 100 abstract nouns, 109 adverbs, and 100 verbs, covering a broad range of semantic types. Detailed statistics are provided in Appendix~\ref{appendix:details}, and the full dataset is available in our repository. Based on preliminary experiments, we found that concrete noun pairs yield the most stable and interpretable game dynamics. Accordingly, our main evaluation uses 464 game instances across twelve categories: \textit{food, landforms, animals, artifacts, tools, people/social, plants, sports, stationer, electronics, clothing}, and \textit{sundries}.
CK-Arena is designed to be extensible. Users can construct domain-specific concept sets for specialized evaluations, and Appendix~\ref{appendix:scalable} provides practical guidelines and examples for extending the benchmark.

\paragraph{Evaluation Metrics.}
\label{metrics}
CK-Arena evaluates model performance using two complementary categories of metrics.
\textbf{(A) Player-level metrics} capture overall game outcomes. \textit{Win Rate (WR)} measures the proportion of games a model wins, reflecting its ability to fulfill role objectives. \textit{Survival Rate (SR)} measures how long a model remains in the game before elimination, indicating its capability to navigate social dynamics and avoid suspicion.
\textbf{(B) Statement-level metrics} assess the quality of individual responses during gameplay. Both are normalized to a 0–1 scale. \textit{Novelty} measures how much new information a statement contributes relative to previous ones, discouraging repetition and encouraging informative descriptions. \textit{Reasonableness} evaluates whether a statement is logically consistent with the underlying concept. Statements failing either criterion are automatically filtered to maintain meaningful gameplay.

\paragraph{Large Language Models as Judges.}

To meet the extensive knowledge demands of diverse topics, we adopt a multi-judge pipeline: strong LLMs from different families first produce independent assessments using prompts aligned with the evaluation framework in Section~\ref{metrics}, where each dimension is defined together with scoring rubrics and worked examples to ensure round-by-round consistency. In order to prevent instability caused by LLMs as judgments, we have set up a manual team to review and adjust some scores based on LLMs' analysis process and relevant open source knowledge bases \cite{miller1995wordnet,wikidata2025,wikipedia2025}. Specifically, 3.1\% of the scores were manually calibrated. Once a sufficient volume of annotated data has been collected, the judging process can be further automated, as described below.

\paragraph{Result Collection and Analysis.}
CK-Arena records comprehensive information throughout each game session. For every game, the system generates a structured JSON log containing metadata such as game ID, timestamp, and concept pairs, along with player information including model identity, assigned role, and concept. It also records all player statements with their corresponding novelty and reasonableness scores, voting outcomes, elimination results, and overall game statistics.
Users may choose to retain either full reasoning traces or only statements and voting records, depending on analysis needs. All data are organized by game rounds, enabling fine-grained analysis of interaction dynamics and decision-making behavior. Moreover, automated scripts aggregate results across games to produce statistical summaries and visualizations, including decision accuracy, elimination patterns, and statement-level metrics.


\paragraph{Unified Rating System.} 
To enable consistent evaluation across repeated experiments, we introduce a unified rating system that tracks model performance over multiple games in CK-Arena. Since player behavior involves multiple aspects such as winning, survival, and voting accuracy, we adopt a composite evaluation scheme rather than relying on a single metric.
Specifically, we employ a team-based Elo rating system adapted to the CK-Arena setting. Each player’s rating is updated dynamically based on game outcomes, individual performance, opponent strength, and experience-dependent uncertainty~\citep{elo1978rating}. For player~$i$ in game $g$, we define a composite performance score
$S_i^g = \alpha \cdot W_i^g + \beta \cdot SR_i^g + \gamma \cdot VR_i^g$,
where $W_i^g \in {0,1}$ denotes the win outcome, $SR_i^g \in [0,1]$ represents the survival rate, and $VR_i^g \in [0,1]$ denotes the voting accuracy. In all experiments, we set $(\alpha, \beta, \gamma) = (0.75, 0.15, 0.10)$.

To account for uncertainty differences between early and late evaluations, we apply an experience-dependent K-factor that decays across batches rather than individual games. Games are grouped into batches of twelve to reflect topic-level variation and to stabilize updates across diverse concept sets. The K-factor is defined as
$K(n) = K_{min} + (K_{max} - K_{min}) \cdot \exp\left(-\frac{\lfloor n/12 \rfloor}{\tau}\right)$,
where $n$ represents the number of games played and we set $K_{max} = 60$, $K_{min} = 5$, $\tau = 2.5$. This formulation ensures high volatility for new players ($K \approx 60$ at $n=0$) while stabilizing ratings for experienced players ($K \approx 5$ at $n \geq 140$).

During preliminary analysis, we observed a consistent role bias in the Undercover game. Under the two-versus-four setting, civilians win more frequently, with an average win rate of approximately $66.7\%$. Similar observations have been reported in prior studies~\citep{dong2024undercover,xu-zhong-2025-comet}. To correct for this imbalance, we introduce a temporary Elo offset of $+120$ for the civilian role when computing expected performance. This adjustment ensures that players with comparable skill levels receive fair rating updates regardless of role assignment. A detailed sensitivity analysis of this offset are provided in Appendix~\ref{appendix:120}.



\subsection{Snapshot QA Benchmark}\label{static}
\paragraph{Generated Data for QA Benchmark.}
CK-Arena is designed to evaluate conceptual understanding beyond win–loss outcomes. Relying solely on game results is insufficient to fully characterize a model’s conceptual ability. A key advantage of CK-Arena is that, during gameplay, LLM agents naturally produce rich, multi-level descriptions of concepts and their distinguishing features. These interaction traces can be systematically reused to construct a high-quality \textit{Question Answering (QA) benchmark}.

From 500 completed games, we automatically extract high-quality statements and generate a total of 5733 QA instances. These instances are organized into three task types that probe different aspects of conceptual understanding:
\textbf{(A) Fine-Grained Comparison:} Given a concept pair and a description, determine which concept the description best matches.
\textbf{(B) Cross-Concept Inference:} Given a concept and an opponent’s description, infer the most likely related concept.
\textbf{(C) Outlier Statement Detection:} Given a concept and four statements, identify the one that does not describe the target concept.

All questions are associated with clear ground-truth labels derived directly from gameplay logs, enabling reliable and large-scale evaluation without manual annotation.

\paragraph{Advantages of the Snapshot Benchmark.}
The snapshot benchmark complements the interactive evaluation by decomposing complex gameplay behavior into well-defined diagnostic tasks. This allows researchers to analyze specific strengths and weaknesses of LLMs, rather than relying solely on aggregate game performance.

Compared to traditional QA benchmarks that require costly manual data collection and are prone to overfitting, our snapshot benchmark is generated automatically from gameplay traces. This makes it scalable, inexpensive to expand, and robust to memorization effects. Because the data originate from diverse game interactions rather than fixed question templates, the benchmark remains dynamic and resistant to overfitting, while still providing controlled, reproducible evaluation signals.
Together, the dynamic game evaluation and the derived snapshot benchmark form a complementary framework that enables both holistic and fine-grained assessment of conceptual understanding.

\begin{table}[t]
\centering
\caption{Performance comparison in CK-Arena. Results are reported separately for the \textit{Civilian} and \textit{Undercover}. WR (\textit{Win Rate}) and SR (\textit{Survival Rate}) serve as indicators of in-game performance, where higher values reflect stronger capability in fulfilling role objectives. 
Reasonableness (Reason.) measures the logical consistency of statements with the target concept, while Novelty evaluates the degree of new information introduced. 
We show how models balance these factors, with Qwen2.5-72b leading in reasonableness, GPT-4o showing strong civilian win rates, and Gemini-2.0-pro-exp excelling in novelty. 
The best values are in \textbf{bold} and the second-best are \underline{underlined}.}
\label{tab:llm_comparison}
\scriptsize    
\setlength{\tabcolsep}{5pt}
\renewcommand{\arraystretch}{1.15}
\begin{tabular}{@{}llcccc@{}}
\toprule
\multirow{2}{*}{\textbf{LLM}} & \multirow{2}{*}{\textbf{Role}} & \multicolumn{4}{c}{\textbf{Performance Metrics}} \\
\cmidrule(l){3-6}
& & $\textbf{WR} \uparrow$ & $\textbf{SR} \uparrow$ & $\textbf{Reason.} \uparrow$ & $\textbf{Novelty} \uparrow$\\ 
\midrule
\multirow{2}{*}{Qwen2.5-72b} 
& Civilian   & \underline{0.6847} & \textbf{0.7207} & 0.9593 & 0.6676 \\ 
& Undercover & \textbf{0.3636} & \textbf{0.2955} & \textbf{0.9737} & 0.7051 \\ 
\midrule
\multirow{2}{*}{GPT-4o} 
& Civilian   & \textbf{0.6854} & 0.6629 & \underline{0.9678} & 0.6693 \\ 
& Undercover & 0.3485 & 0.2273 & 0.9614 & 0.7429 \\ 
\midrule
\multirow{2}{*}{DeepSeek-v3} 
& Civilian   & 0.6814 & \underline{0.6637} & 0.9470 & 0.8248 \\ 
& Undercover & \underline{0.3571} & 0.2857 & 0.9537 & 0.8220 \\ 
\midrule
\multirow{2}{*}{LLaMA-3.3-70b} 
& Civilian   & 0.6702 & 0.6596 & 0.9663 & 0.8072 \\ 
& Undercover & 0.3279 & 0.1803 & 0.9678 & 0.8083 \\ 
\midrule
\multirow{2}{*}{Gemini-2.0-pro-exp} 
& Civilian   & 0.6636 & 0.6545 & 0.9667 & \underline{0.8259} \\ 
& Undercover & 0.3111 & \underline{0.2889} & 0.9652 & \textbf{0.8391} \\ 
\midrule
\multirow{2}{*}{Claude-3-5-Haiku} 
& Civilian   & 0.6408 & 0.6214 & 0.9494 & 0.7633 \\ 
& Undercover & 0.2692 & 0.1923 & 0.9273 & 0.8061 \\ 
\bottomrule
\end{tabular}
\end{table}


\begin{figure}[t] 
    \centering
    \vspace{-0.5cm}
    \includegraphics[width=\columnwidth]{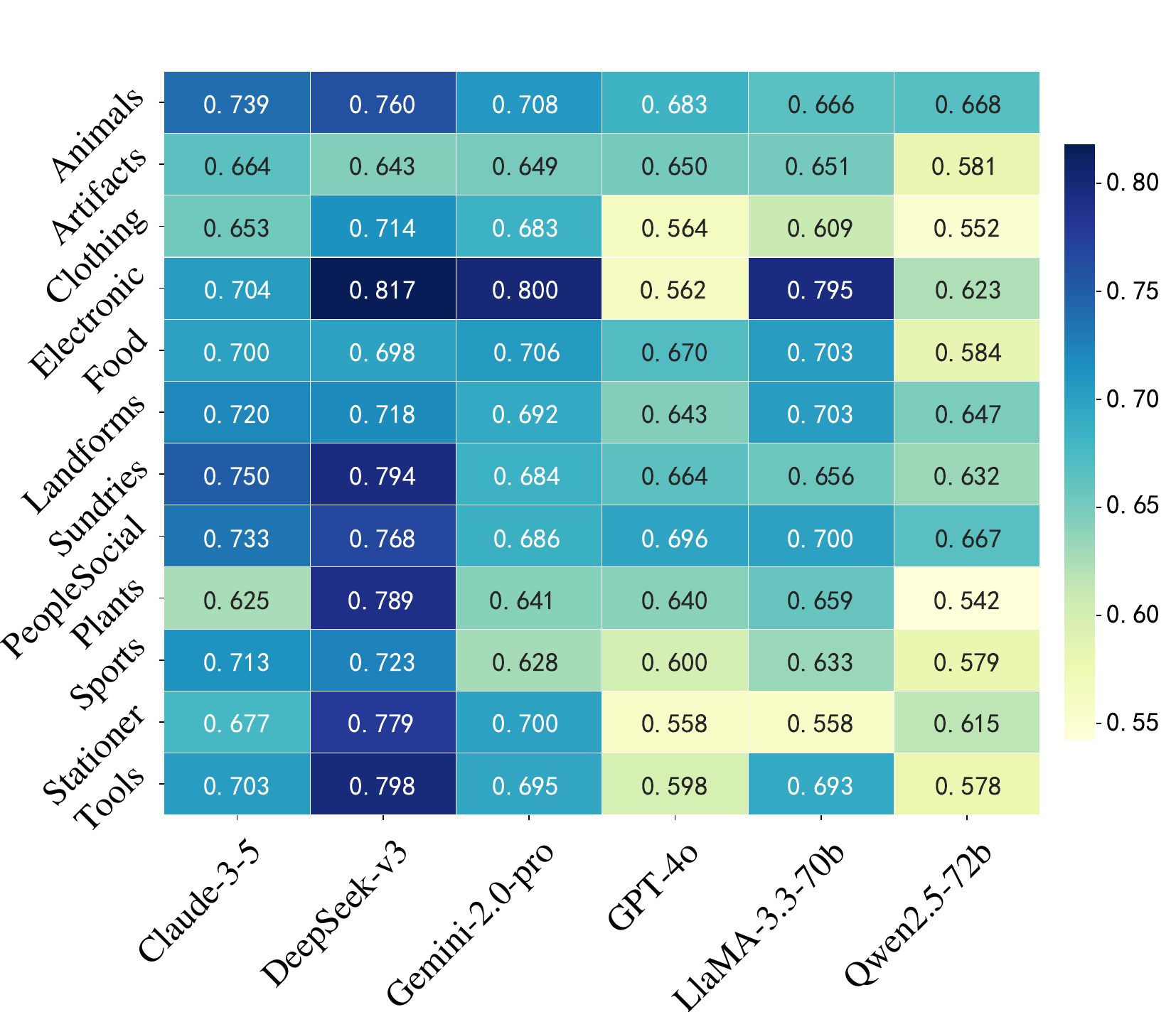}
    \caption{Relevance scores of different LLMs across various categories. 
The heatmap shows how well model statements align with target concepts, 
where darker colors indicate higher relevance.
    }
    \label{fig:heat}
    \vspace{-0.2cm}
\end{figure}

\begin{figure*}[t] 
    \centering
    \begin{subfigure}{0.32\textwidth}
        \includegraphics[width=\textwidth]{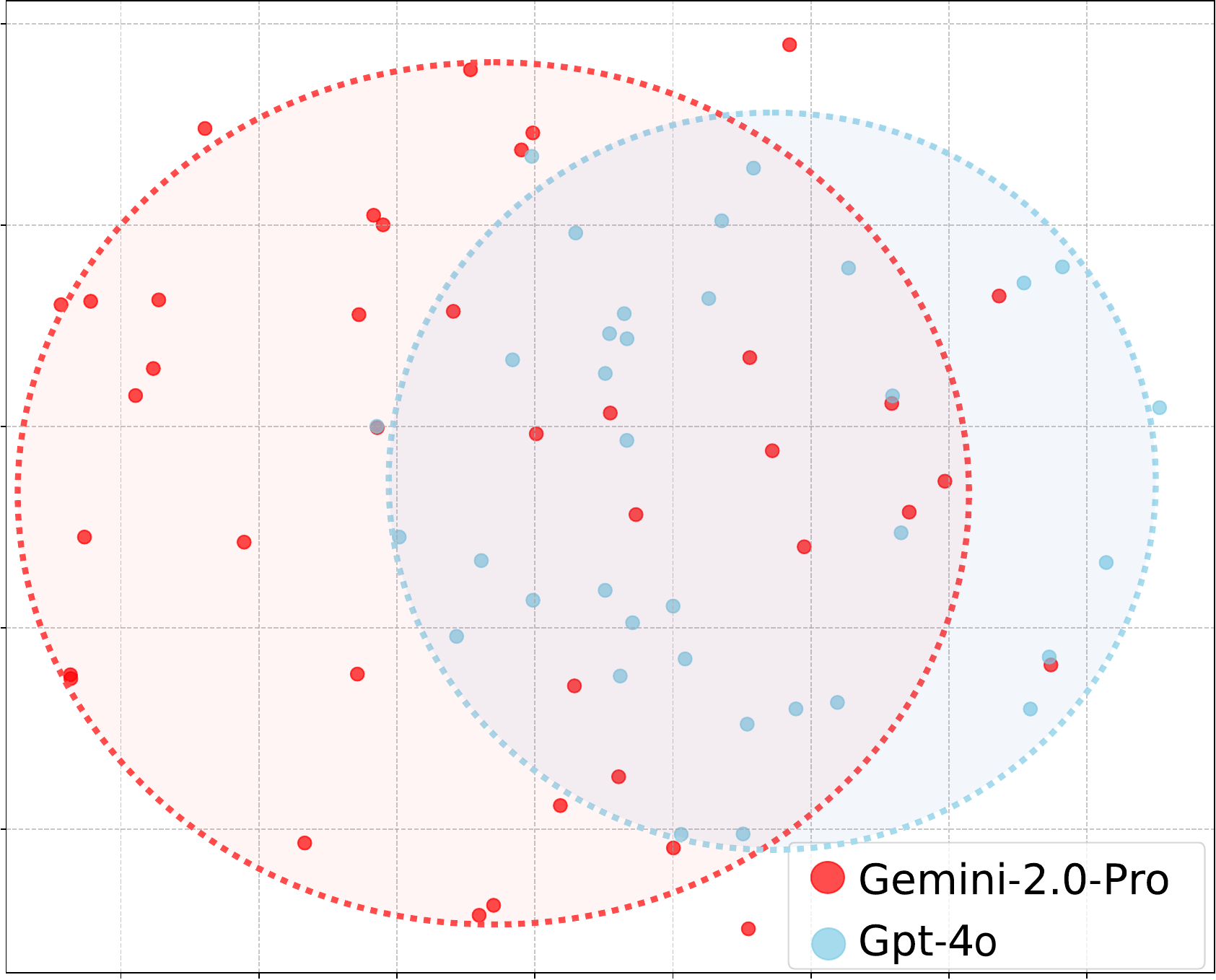}
        \caption{Animals}
    \end{subfigure}
    \begin{subfigure}{0.32\textwidth}
        \includegraphics[width=\textwidth]{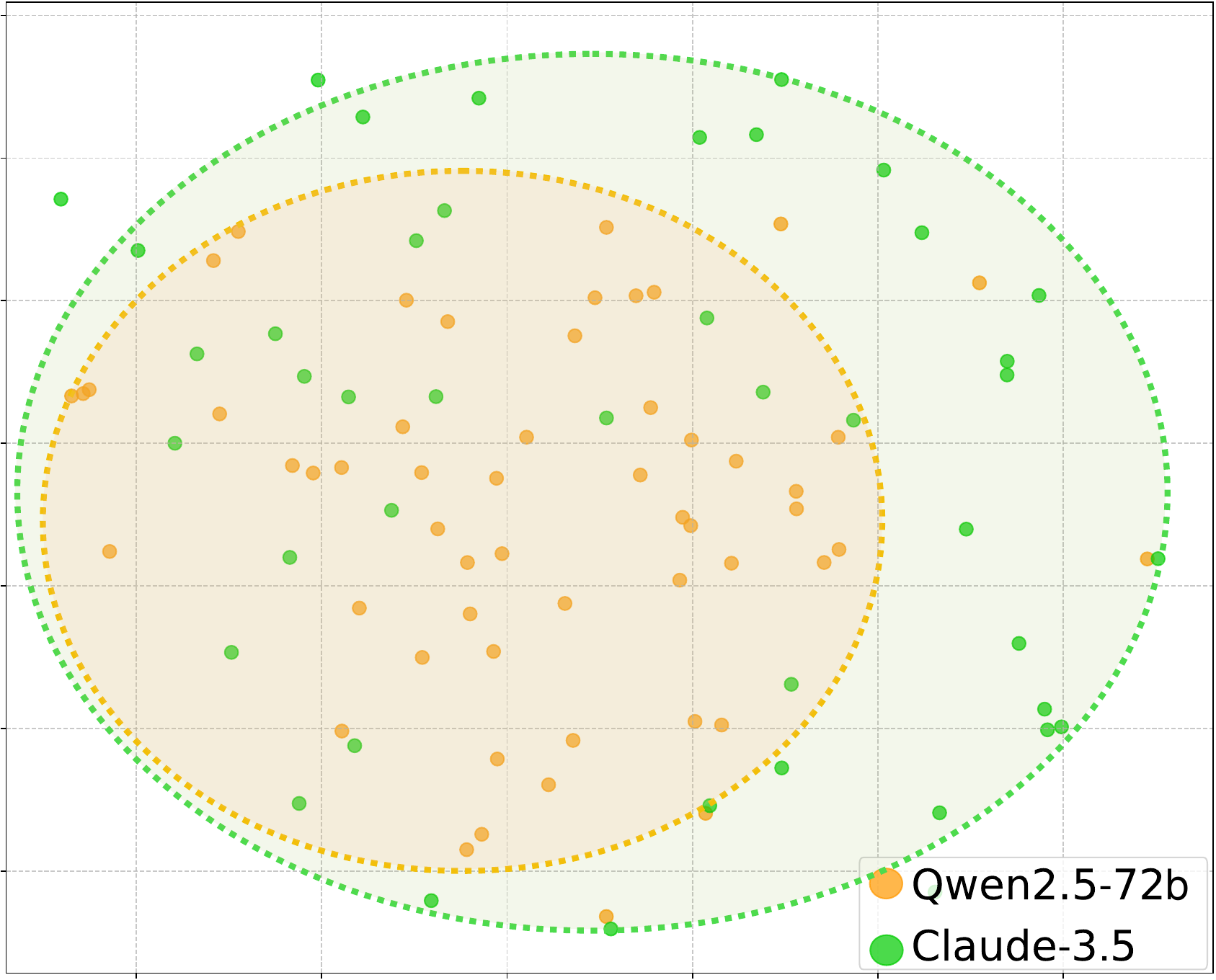}
        \caption{Food}
    \end{subfigure}
    \begin{subfigure}{0.32\textwidth}
        \includegraphics[width=\textwidth]{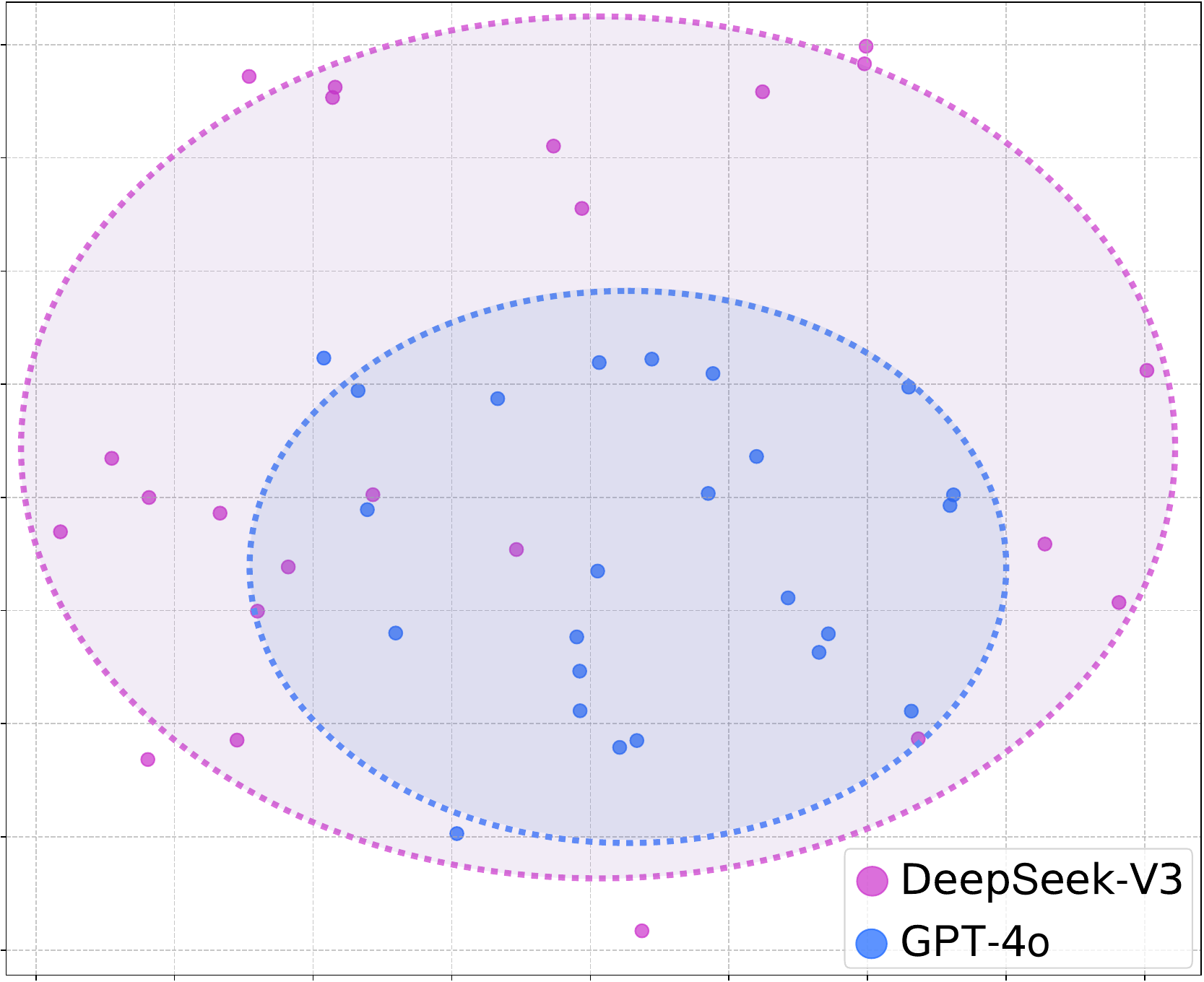}
        \caption{Electronic}
    \end{subfigure}
    \caption{\textbf{t-SNE visualizations of LLM statements across concept categories.} 
    Each plot shows model outputs for (a) Animals, (b) Food, and (c) Electronics. Repetitive descriptions, reflecting shallow understanding, appear as tightly clustered points, whereas richer knowledge produces more dispersed distributions. The visualizations also indicate that different LLMs center their descriptions on different focal aspects of a concept, suggesting variation in how conceptual knowledge is represented.}
    \label{fig:tsne}
\end{figure*}

\section{Experiments}

Our experiments proceed in three stages. \textbf{First}, we evaluate six representative LLMs from different families in controlled six-player games, focusing on statement quality, conceptual understanding, and role-specific performance. \textbf{Second}, we construct a scalable leaderboard in which additional models are compared against these baselines, enabling consistent cross-model evaluation. \textbf{Third}, we convert the interaction traces from gameplay into a snapshot benchmark with fine-grained tasks, allowing us to analyze factors that influence game outcomes and compare model behavior at a more detailed level.
Overall, we conduct 500 games for ranking evaluation and additional games for analysis. These games generate 7412 concept descriptions, from which we construct 5733 QA-style test instances. We further include quantitative analyses and ablation studies to reduce the impact of strategy and ensure that performance reflects conceptual understanding. Additional results are provided in Appendix~\ref{appendix:details} and Appendix~\ref{appendix:moreexp}.

\subsection{Experimental Setup}

We evaluate six widely used LLMs from different families: \textit{Claude-3-5-Haiku}~\citep{anthropic2024claude35}, \textit{GPT-4o}~\citep{hurst2024gpt}, \textit{Gemini-2.0-Pro-Exp}~\citep{team2023gemini}, \textit{DeepSeek-v3}~\citep{liu2024deepseek}, \textit{LLaMA-3.3-70b}~\citep{grattafiori2024llama}, and \textit{Qwen2.5-72b}~\citep{qwen}. For statement-level evaluation, \textit{GPT-4.1-2025-04-14}~\citep{openai2025gpt41} and \textit{Claude-3-7-Sonnet-20250219}~\citep{anthropic2024claude37} serve as LLM-based judges to score all statements. A human expert panel then reviewed all statements, taking into account both the LLM scores and relevant reference knowledge, to determine the final scores.

For leaderboard construction, \textit{DeepSeek-v3} and \textit{Qwen2.5-72b} serve as anchor models. We evaluate additional LLMs including \textit{GPT-5}~\citep{openai2025gpt5}, \textit{GPT-oss-120b}~\citep{openai2025gptoss120bgptoss20bmodel}, \textit{DeepSeek-v3.1}~\citep{deepseek3.1}, \textit{Claude-opus-4.1}~\citep{anthropic2025claude41}, \textit{Claude-opus-4.5}~\citep{anthropic2025claude45}, \textit{Kimi-k2-instruct}~\citep{kimiteam2025kimik2openagentic}, \textit{Qwen-plus}~\citep{qwen3}, \textit{Qwen3-max}~\citep{qwen3max}, \textit{Ernie-4.5-300b-a47b}~\citep{ernie2025technicalreport}, \textit{Gemini-2.5-flash-preview}~\citep{gemini2.5flash}, and \textit{Gemini-3-pro-preview}~\citep{gemini3}. Each model receives identical prompts and plays at least 60 rounds against the anchor models to ensure rating stability.

\subsection{Concept Understanding Evaluation}
\label{results}

\paragraph{Performance Comparison.} Table~\ref{tab:llm_comparison} summarizes baseline performance in CK-Arena. Across 6 LLMs, civilian win rates are consistently higher than undercover win rates, reflecting the greater difficulty of the undercover role, which requires concealing one’s concept while inferring shared features with the civilian concept.
All models achieve high reasonableness scores. We speculate that this is partly due to the threshold-based filtering mechanism, which removes clearly invalid responses before evaluation. The results also indicate that modern LLMs can reliably follow instructions and generate logically coherent descriptions grounded in basic conceptual knowledge.

Novelty exhibits more nuanced behavior. Strong models such as \textit{Qwen2.5} and \textit{GPT-4o} often score lower on novelty because overly creative descriptions increase the risk of exposing the undercover identity, while overly repetitive statements are penalized. Effective performance therefore requires balancing informativeness and caution. CK-Arena captures this trade-off, making it a meaningful test of a model’s ability to express fine-grained conceptual distinctions rather than simply generating diverse or safe responses.

\paragraph{Statement–Concept Relevance.}
We further evaluate \textit{Relevance}, which measures how well each statement aligns with the assigned concept. Higher scores indicate more specific and concept-aligned descriptions that help civilians identify the undercover agent, while lower scores correspond to broader or more ambiguous statements that could apply to multiple concepts. This metric reflects an inherent tension in the game: civilians benefit from precise descriptions, whereas undercover agents may intentionally remain vague to avoid exposure.
Figure~\ref{fig:heat} presents relevance scores across conceptual categories. Models such as \textit{DeepSeek-v3} achieve higher relevance scores, while others such as \textit{Qwen2.5-72b} score lower, yet both obtain strong win rates, showing that relevance alone does not predict overall performance.

\paragraph{Semantic Dispersion as a Proxy for Conceptual Depth.}
We embed LLM-generated statements and compare them using dimensionality reduction and visualization. Given the same number of descriptions for a concept, shallow understanding typically leads to repetitive phrasing, which appears as tightly clustered points in the t-SNE plot, whereas deeper knowledge produces more dispersed patterns. The visualizations reveal systematic differences in how models generate conceptual descriptions under the same topic. Figure~\ref{fig:tsne}(a) shows that \textit{Gemini-2.0-pro-exp} and \textit{GPT-4o} emphasize different aspects of the same concept, reflecting variation in conceptual focus. Figures~\ref{fig:tsne}(b) and (c) further demonstrate differences in clustering degree, with some models producing narrow clusters and others spreading more broadly across the semantic space. This indicates the variation in focus and the degree of dispersion in LLM-generated conceptual associations.


\subsection{Model Ranking via CK-Arena Leaderboard}

\begin{figure}[!t]
        \includegraphics[width=\columnwidth]{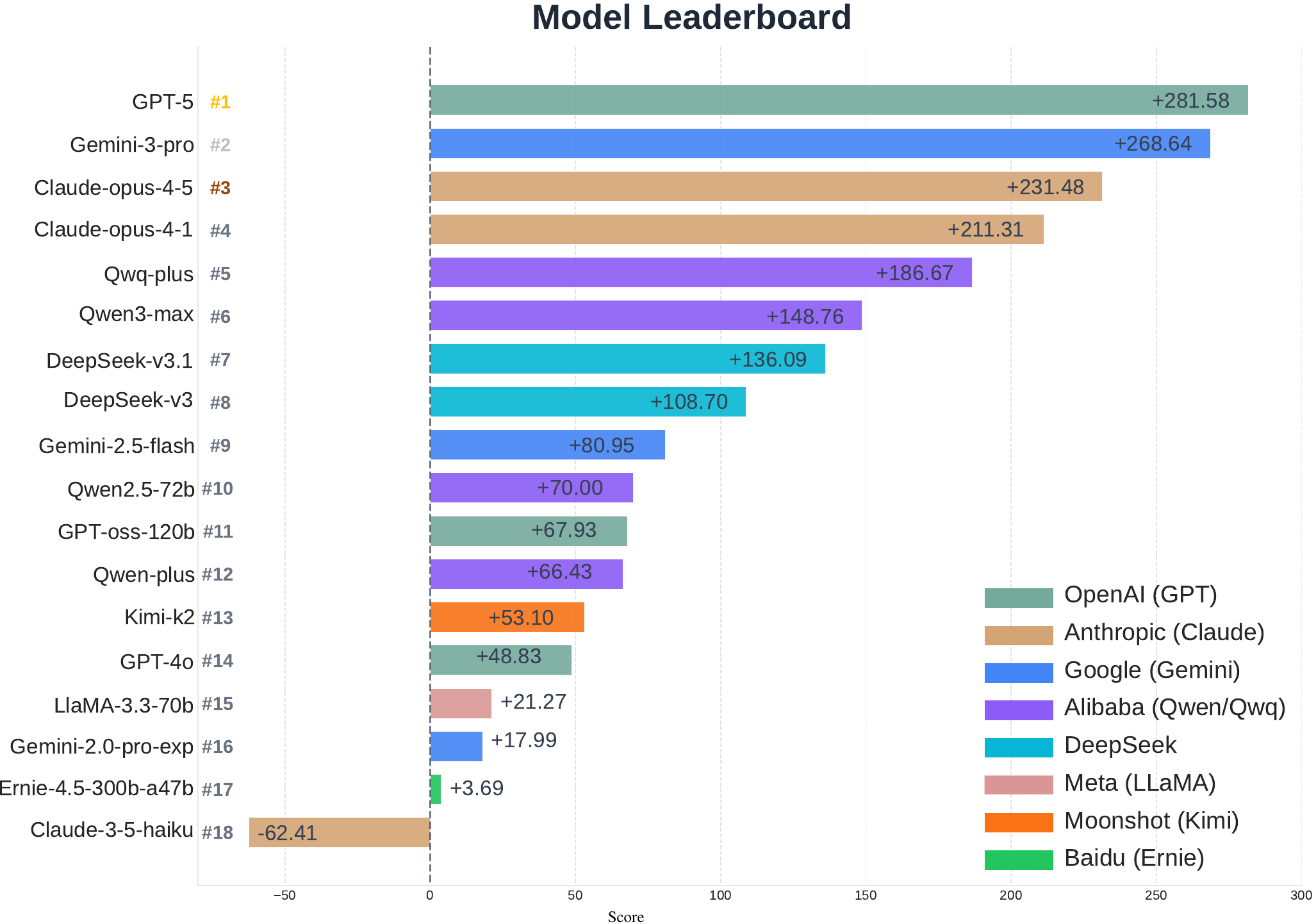}
        \caption{Leaderboard of LLMs in CK-Arena.Each player starts with an initial rating of 0. After stabilization, a player consistently defeating 0-rated opponents converges around 420, which serves as a reference for strong performance. The leaderboard highlights relative differences across 18 evaluated LLMs.}
        \label{fig:elo}
\end{figure}

Using the unified rating system, we construct a leaderboard covering 18 LLMs (Figure~\ref{fig:elo}). All models start with an initial rating of 0. After convergence, a model that consistently defeats baseline opponents stabilizes around an Elo score of 420, which serves as a reference level for strong performance. To ensure robustness, we evaluate models under both forward and reverse ordering. The two settings produce nearly identical rankings, with a maximum Elo difference of 1.72 and a Pearson correlation of 0.99, confirming the stability of the evaluation.

We additionally include a human baseline to provide a reference for general conceptual reasoning. However, humans face inherent disadvantages in this setting, as CK-Arena requires broad and consistent knowledge across diverse domains, which is difficult for individuals to maintain. Further details are provided in Appendix~\ref{appendix:moreexp}.

Within model families such as \textit{DeepSeek} and \textit{Qwen}, performance differences between newer and older versions are relatively modest, and all remain behind top-tier models such as \textit{GPT-5}. This suggests that incremental model updates alone are insufficient to close the performance gap, a distinction that CK-Arena is able to reveal clearly.

\subsection{QA Benchmark for Fine-Grained Diagnosis}

\begin{figure}[t]
        \includegraphics[width=\columnwidth]{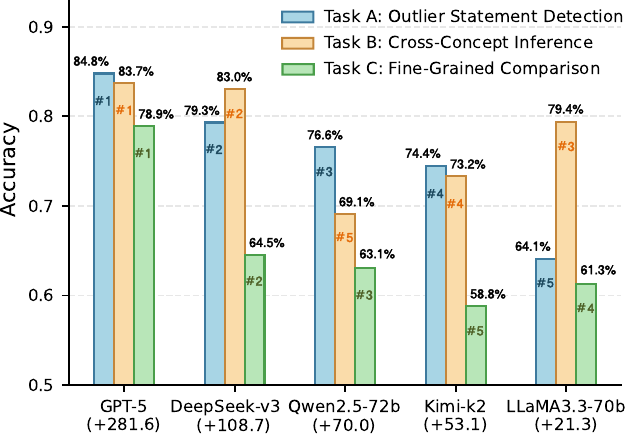}
        \caption{QA benchmark results. Three open-source models were selected for this evaluation, and the results reflected the specific performance of different models on a single task. The consistency between QA benchmark ranking and dynamic game ranking can also indirectly demonstrate the reliability of dynamic evaluation.}
        \label{fig:qa}
\end{figure}

We further evaluate models using the QA benchmark described in \cref{static} to analyze conceptual understanding at a finer level. The benchmark is constructed from gameplay traces and targets specific reasoning skills rather than overall game success. We evaluate a diverse set of models, including the top-performing model on the CK-Arena leaderboard and several representative open-source LLMs, covering different model families and capability levels. Results in \cref{fig:qa} show a strong correlation between QA accuracy and game win rate (mean Spearman $\rho = 0.87$), indicating that CK-Arena performance reflects genuine conceptual understanding rather than game-specific strategies. Moreover, task-level results reveal why models differ in performance. For example, \textit{GPT-5} achieves higher accuracy on cross-concept inference, suggesting stronger ability to organize and reason over conceptual features, which directly explains its advantage in the game. These findings confirm that CK-Arena captures meaningful differences in conceptual reasoning, and that the QA benchmark provides an interpretable complement to the game-based evaluation.

\subsection{Ablation Study: Strategy \textit{vs.}\ Knowledge Mastery}

\begin{table}[t]
\centering
\caption{Win-rate comparison of Qwen2.5-72b-Instruct under different game strategies. Each batch of games chooses different restriction strategies and faces the same baseline as Sec. \ref{results}.}
\label{tab:ablation}
\newcommand{\boldbest}[1]{\textbf{#1}}   
\newcommand{\subopt}[1]{\underline{#1}}  
\resizebox{\linewidth}{!}{%
\begin{tabular}{lcccc}
\toprule
\textbf{Model} & \textbf{Strategy} & \textbf{Civ. WR $\uparrow$} & \textbf{Und. WR $\uparrow$} & \textbf{Ov. WR $\uparrow$} \\
\midrule
Qwen2.5-72b-Instruct & Normal      & \boldbest{0.685} & \boldbest{0.364} & \boldbest{0.594} \\
Qwen2.5-72b-Instruct & Conservative& 0.517 & \subopt{0.357} & 0.465 \\
Qwen2.5-72b-Instruct & Aggressive  & \subopt{0.664} & 0.196 & 0.506 \\
Qwen2.5-72b-Instruct & Random & 0.612 & 0.265 & \subopt{0.527} \\
\bottomrule
\end{tabular}%
}
\vspace{-0.5cm}
\end{table}

\begin{table}[t]
\centering
\caption{
\textbf{Performance of GPT-5 under different fixed strategies in the same game.} Fixed strategies (random, specific, vague) consistently underperform the unconstrained setting, indicating that GPT-5’s strong performance does not stem from a single universal strategy. Results for DeepSeek-v3 and Qwen2.5-72b are included for reference.
}
\label{tab:ablation_gpt5}
\resizebox{\linewidth}{!}{%
\begin{tabular}{lcccccc}
\toprule
\multirow{2}{*}{\textbf{Model}}
& \multicolumn{4}{c}{\textbf{GPT-5}}
& \textbf{Deepseek-v3}
& \textbf{Qwen2.5-72b} \\
\cmidrule(lr){2-5}\cmidrule(lr){6-6}\cmidrule(lr){7-7}
& Normal & Random & Aggressive & Conservative & Normal & Normal \\
\midrule
\textbf{Score}
& +170 & +132 & +57 & -16 & +10 & -5 \\
\bottomrule
\end{tabular}%
}
\end{table}

A potential concern is that models might achieve high win rates through superficial strategies rather than genuine conceptual understanding. To examine this, we conduct ablation experiments in which models are forced to follow fixed strategies during gameplay: 1) Conservative: always produce vague, high-level descriptions; 2) Aggressive: always generate highly specific descriptions; 3) Random: alternate randomly between conservative and aggressive behaviors.

We evaluate GPT-5, which performs strongly on the leaderboard, and the open-source model Qwen2.5-72b. As shown in \cref{tab:ablation}, all fixed-strategy settings lead to substantial performance drop compared to the unconstrained baseline, indicating that heuristic behavior alone is insufficient.

We further conduct a controlled experiment using GPT-5 only (\cref{tab:ablation_gpt5}), where six agents compete in the same game: four GPT-5 agents following different strategies (three fixed strategies and one unconstrained) and two anchor players for calibration. This setup enables direct comparison among strategies under identical conditions. The results consistently show that the unconstrained GPT-5 agent outperforms all fixed-strategy variants.

Across both experiments, no single strategy yields consistently strong performance. Instead, success in CK-Arena requires adaptive reasoning grounded in conceptual understanding, including the ability to adjust description specificity based on role, context, and other players’ statements.

\section{Conclusion}
We introduce CK-Arena, a dynamic benchmark for evaluating conceptual understanding in large language models through multi-agent interaction. Built on the \textit{Undercover} game, it provides a scalable and dynamic environment where models engage with associations, similarities, and differences between concepts, an ability that traditional static benchmarks often overlook.  Experiments show that conceptual understanding varies across categories and does not consistently align with general benchmark performance, indicating that skills such as coding or mathematics do not necessarily translate into stronger conceptual understanding. 
We hope this benchmark serves as a useful foundation for future work on evaluating and improving the semantic reasoning abilities of LLMs.
%

\section*{Impact Statement}
This paper presents work whose goal is to advance the field of Machine Learning, specifically in the area of LLM evaluation methodology. 
It establishes a game-based dynamic evaluation benchmark, which has been validated in strategy games, but needs to be attempted in games involving knowledge. 
Compared to static benchmarks, dynamic benchmarks have advantages such as being less prone to data pollution, relying on rules rather than manually annotating data, being closer to real-world scenarios, and examining comprehensive capabilities. They are of great significance for the development of large language models.

\bibliography{ref}
\bibliographystyle{icml2026}

\newpage
\appendix
\onecolumn

\section{Implementation Details} \label{appendix:details}

\paragraph{Detailed Data Statistics}
The dataset we provided contains a total of 529 English pairs of concepts, including 220 concrete noun pairs, 100 abstract noun pairs, 109 adverb pairs, and 100 verb pairs. After initial experimental attempts, we concluded that concrete noun pairs are more suitable for our experimental setup and overall research questions. Therefore, for the specific experiments, we selected 12 different categories from the 220 concrete noun pairs. These categories consist of concrete noun pairs that are closest to our daily life and conversational contexts. All of those concepts can be considered with rich and clearly describable features.
We believe that starting with these concept pairs can more reliably and steadily complete our experiments and yield preliminary results. In the future, we will further explore the other words.

\paragraph{QA Snapshot Benchmark Construction.}
We construct the QA snapshot benchmark through a two-stage pipeline. First, we automatically extract all player utterances from gameplay logs, together with metadata such as game outcomes, voting results, and associated concepts. Second, we apply task-specific filtering rules based on in-game behavior and judge scores to ensure data quality. For Task A (Fine-Grained Comparison) and Task B (Cross-Concept Inference), we retain only statements with a relevance score $\geq 0.8$ and a reasonableness score $\geq 0.9$. For Task C (Outlier Statement Detection), we adopt a behavior-driven criterion, selecting statements that led to the speaker being voted out in the corresponding round. The entire pipeline is fully automated, with optional manual refinement to support customized evaluation settings.

\paragraph{Connection of CK-Arena with Existing Benchmarks.}
We think that CK-Arena offers distinct yet complementary value in LLM evaluation tasks.
Fundamental Differences in Evaluation Focus: Traditional benchmarks like MMLU primarily assess factual recall and static knowledge retrieval through multiple-choice questions. In contrast, CK-Arena evaluates dynamic conceptual understanding in interactive contexts. For example, while MMLU might ask ``Which of the following animals is a primate?'', CK-Arena requires models to articulate the distinguishing features between closely related concepts (e.g., monkeys vs. apes) and navigate the semantic boundaries dynamically based on partial information from other agents.

Why Static vs. Dynamic Evaluation Matters: Our preliminary analysis suggests that strong performance on traditional benchmarks doesn't necessarily translate to effective conceptual boundary navigation. For instance, a model might correctly identify that both soccer and basketball are sports (factual knowledge) but struggle to strategically describe one while concealing its identity when the other is the majority concept (conceptual understanding + strategic reasoning). This highlights that knowing facts about concepts differs from understanding their relational structures and boundaries.

We also point out that CK-Arena does not aim to replace existing benchmarks but to fill a critical gap in evaluating interactive conceptual understanding. Traditional benchmarks excel at measuring breadth of knowledge, while CK-Arena probes depth of conceptual understanding in realistic social contexts. The differences in results reflect that multi-agent interaction requires different cognitive processes than isolated question-answering.

\paragraph{Scalability Demonstration.}

In order to explain the scalability of CK-Arena, we provide a specific example in this section to help researchers who need to build their own datasets test LLM's knowledge mastery in specific fields. We will divide this task into three steps:

Firstly, researchers need to construct concept pairs related to evaluation knowledge within their field (for example, by describing the similarities and differences between alcohol lamps and flame spray guns to explore the knowledge of middle school chemistry experiments). Users may also need to adjust prompts if they wish to have their own rating criteria.
Then, users need to conduct at least 60 pre-experiments using models with comparable performance (or one model as all players) and game settings of their own choice (such as number of players, rounds, etc.) to obtain role bias calibration values. Specifically, the concepts in the newly constructed dataset may have inconsistent similarities, which can lead to role bias in the game. For example, if two concepts are very similar, it is obvious that undercover characters are easily mixed up with civilian characters; On the contrary, the undercover character finds it difficult to move forward. Therefore, it is necessary to determine role bias through pre-experiments and use temporary scores to balance this bias.
The third step is for users to repeat the game multiple times until the K-value stabilizes, in order to obtain a performance analysis among the LLM players participating in the game.

Then, here comes the example. Due to the fact that most concepts that contain broadly descriptive features are nouns, our specially designed prompt template is not suitable for evaluating verbs or other parts of speech. Therefore, we carried out a complete extension proces. First, we built a verb word pair dataset, and then adjusted part of the content in the prompt to help players better participate in the game, and judges more standardized. The following are the added parts:

\textit{- Nature of the action: Such as the type characteristics of the action.
- Relationship of action: Such as the characteristics of the subject and object involved.
- Usage scenarios: Such as the environmental characteristics and cultural background where the action occurs.
- Concluding effects: The consequences and impacts brought about by the action.
- Emotional impact: The emotional overtones, moral implications, and social attributes involved in the action.
}

The experimental results regarding verbs can be viewed in section \ref{appendix:moreexp}. During our testing, the API call cost for reviewing a single game was approximately \$0.8, while completing a full theme review required around \$40–50. By replacing expensive LLM-based judges with a fine-tuned model, as mentioned in the paper, these costs could be more substantially reduced. In terms of time efficiency, the open-source code provided in this work already supports batch execution of multiple games. Although API calls impose certain speed constraints, our experiments show that running 5 games in parallel does not trigger rate-limit restrictions, allowing most reviews to be completed in only one day.

\paragraph{Derivation of the 120-Point Elo Offset.} \label{appendix:120}

In this paragraph, we derive the $120$-point Elo offset used to balance the expected performance between the civilian and undercover roles in the game. The goal is to ensure that players of equal skill levels have comparable rating update opportunities, regardless of their assigned roles.

In the Elo rating system, the expected score $E_A$ of player $A$ against player $B$ is given by:
\[
E_A = \frac{1}{1 + 10^{(R_B - R_A)/400}},
\]
where $R_A$ and $R_B$ are the ratings of players $A$ and $B$. Although this is a 1v1 formula, in our design, the Elo update first computes the expected outcome based on the win–loss relationship between two teams, and then incorporates each player’s individual performance for the actual score adjustment. Therefore, we can treat the two teams as player A and player B, and use the standard formula for derivation.

Empirically, the civilian role has a natural advantage, leading to a baseline win probability of $2/3$ for the civilians against undercover agents of equal skill. To determine the Elo offset that corresponds to this advantage, we solve for the Elo rating difference $x$ that yields an expected score of $2/3$:

\begin{equation}
\frac{1}{1 + 10^{-x/400}} = \frac{2}{3}
\end{equation}
\begin{equation}
10^{-x/400} = \frac{1}{2}
\end{equation}
\begin{equation}
-\frac{x}{400} = \log_{10}\!\left(\tfrac{1}{2}\right)
\end{equation}
\begin{equation}
x = 400 \cdot \log_{10}(2) \approx 400 \cdot 0.3010 \approx 120
\end{equation}

Thus, an Elo difference of approximately $120$ corresponds to the observed $2/3$ win rate. 
To balance the game, we introduce a temporary offset of $+120$ Elo points to the civilian side when computing expected outcomes. This adjustment ensures that, from the model’s perspective, the expected probability of winning for both sides is effectively $1/2$, thereby eliminating the systematic role-induced imbalance in rating updates.

\section{More Experimental Results}
\label{appendix:moreexp}

\paragraph{The stability of the scoring process}
To verify the stability of the scoring process in our LLM-based evaluation framework (and thereby support the reliability and repeatability of evaluation results), we conducted three independent evaluations on the animal group. Based on the outcomes of these evaluations, we calculated key statistical indicators, including mean, variance, and standard deviation, for each of the statement-level metrics (Novelty and Reasonableness). The specific statistical data are presented in Table \ref{tab:evaluation_reliability}. This table reflects the stability of the scoring process: LLM-based assessments already demonstrate strong internal consistency, and with additional human review to adjust specific cases, CK-Arena ensures both reproducibility and robustness of the evaluation framework.

\begin{table}[h]
    \centering
    \caption{Statistical indicators of three independent evaluations on the animal group.}
    \label{tab:evaluation_reliability}
    \begin{tabular}{lccc}
        \toprule
        Metric & Mean & Variance & Std Dev \\
        \midrule
        Novelty & 0.8150 & 0.000203 & 0.0142 \\
        Reasonableness & 0.9672 & 0.000042 & 0.0065 \\
        \bottomrule
    \end{tabular}
\end{table}


\begin{figure}[t!]
        \centering
        \includegraphics[width=\linewidth,keepaspectratio]{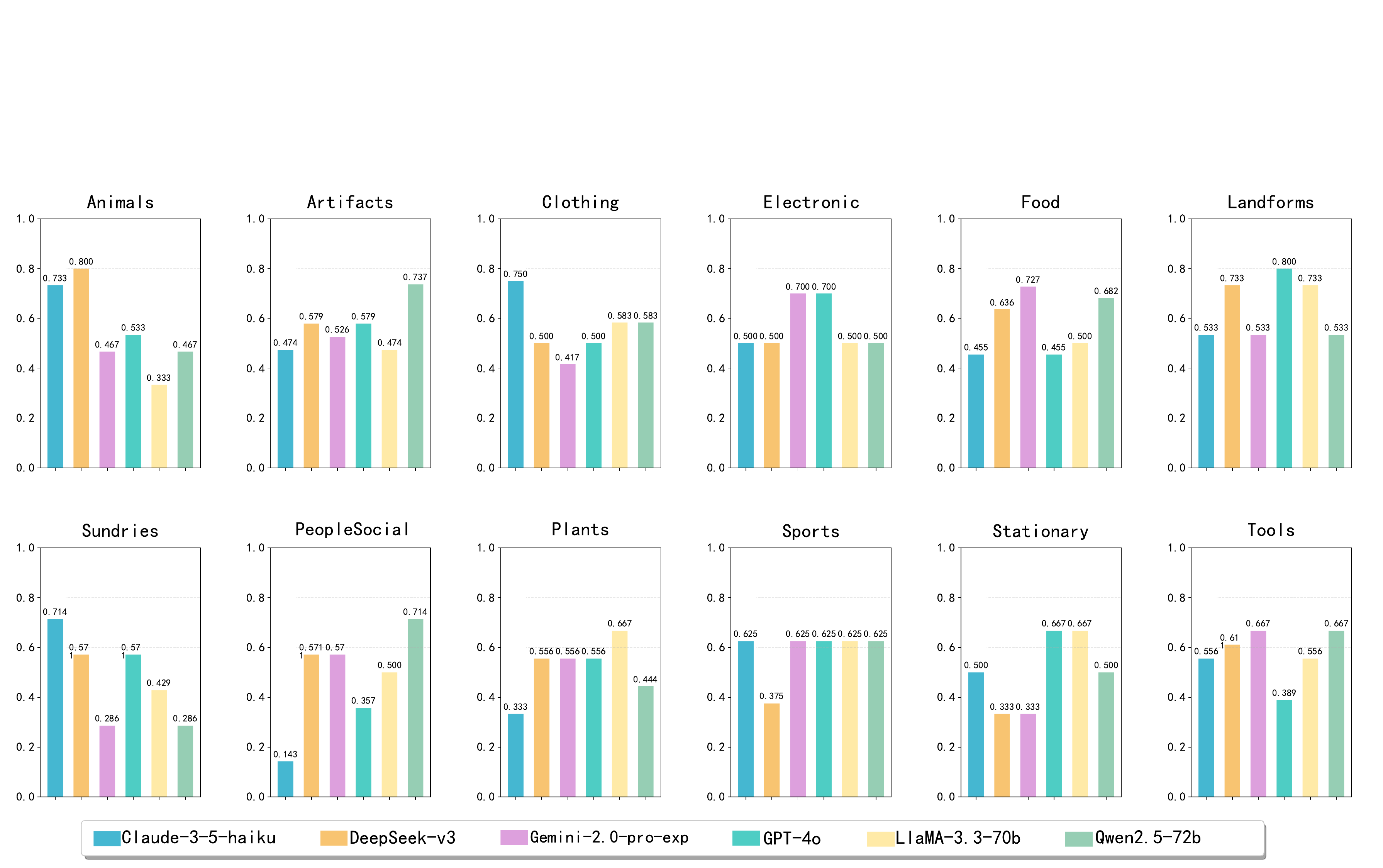}
        \caption{
\textbf{The win rate performance of six LLMs across $12$ categories.} A comparative analysis reveals that each model exhibits distinct strengths and weaknesses across different concept categories. These variations are likely influenced by differences in training data, architectural design, and optimization strategies specific to each model. 
The analysis reveals models’ focus areas, knowledge gaps, and insights for improving conceptual understanding.
        }
        \label{fig:win_rate_combined}
\vspace{-0.2cm}
\end{figure}

\paragraph{Win Rate by Different Categories.}
Figure \ref{fig:win_rate_combined} illustrates the win rate performance of various LLMs across different conceptual categories. The results highlight clear strengths and weaknesses for each model. For example, DeepSeek-v3 achieves the highest win rate in the animal category, reaching $80\%$, indicating strong domain-specific understanding. Similarly, \textit{GPT-4o} excels in the landmark category with a win rate of $80\%$, reflecting its grasp of geographical concepts. In contrast, \textit{Claude-3-5-Haiku} demonstrates a notably low win rate of just $14.3\%$ in the social category, suggesting limitations in handling social context. These performance differences are likely influenced by the models’ training datasets and optimization strategies, highlighting domain-specific expertise and gaps in conceptual understanding.

\begin{wraptable}{r}{0.3\textwidth}
\centering
\vspace{-0.3cm}
\caption{\textbf{Win rate (WR) and Survival rate (SR) comparison of baseline LLMs in CK-Arena with Verbs.} Results are reported separately for \textit{Civilian} and \textit{Undercover} roles.}
\vspace{0.1cm}
\label{tab:llm_wr_sr}
\scriptsize
\setlength{\tabcolsep}{4pt}
\begin{tabular}{@{}llcc@{}}
\toprule
\textbf{LLM} & \textbf{Role} & $\textbf{WR} \uparrow$ & $\textbf{SR} \uparrow$ \\ 
\midrule
\multirow{2}{*}{DeepSeek-v3} 
& Civilian   & 0.429 & 0.739 \\ 
& Undercover & 0.375 & 0.571 \\ 
\midrule
\multirow{2}{*}{Gemini-2.0-exp} 
& Civilian   & 0.476 & 0.429 \\ 
& Undercover & 0.333 & 0.222 \\ 
\midrule
\multirow{2}{*}{Claude-3-5-Haiku} 
& Civilian   & 0.684 & 0.526 \\ 
& Undercover & 0.727 & 0.636 \\ 
\midrule
\multirow{2}{*}{Qwen2.5-72b} 
& Civilian   & 0.565 & 0.739 \\ 
& Undercover & 0.571 & 0.571 \\ 
\midrule
\multirow{2}{*}{LLaMA-3.3-70b} 
& Civilian   & 0.524 & 0.714 \\ 
& Undercover & 0.444 & 0.444 \\ 
\midrule
\multirow{2}{*}{GPT-4o} 
& Civilian   & 0.500 & 0.545 \\ 
& Undercover & 0.375 & 0.375 \\ 
\bottomrule
\end{tabular}
\vspace{-1cm}
\end{wraptable}
\FloatBarrier

\paragraph{Evaluation Based on Verb Vocabulary} \label{appendix:scalable}

We repeated the baseline experiment, but changed the dataset used for evaluation to one with verb themes. Specific data can be found in our open-source repository as shown in Figure \ref{tab:llm_wr_sr}. Interestingly, Claude-3.5, which had always performed at the bottom of the original model, actually achieved the highest win rate in this experiment and showed a significant gap compared to other models. Perhaps we can conduct more fine-grained classification and evaluation to explore the reasons for these phenomena.

\begin{figure*}[t] 
    \centering
    \begin{subfigure}{0.48\textwidth}
        \includegraphics[width=\textwidth]{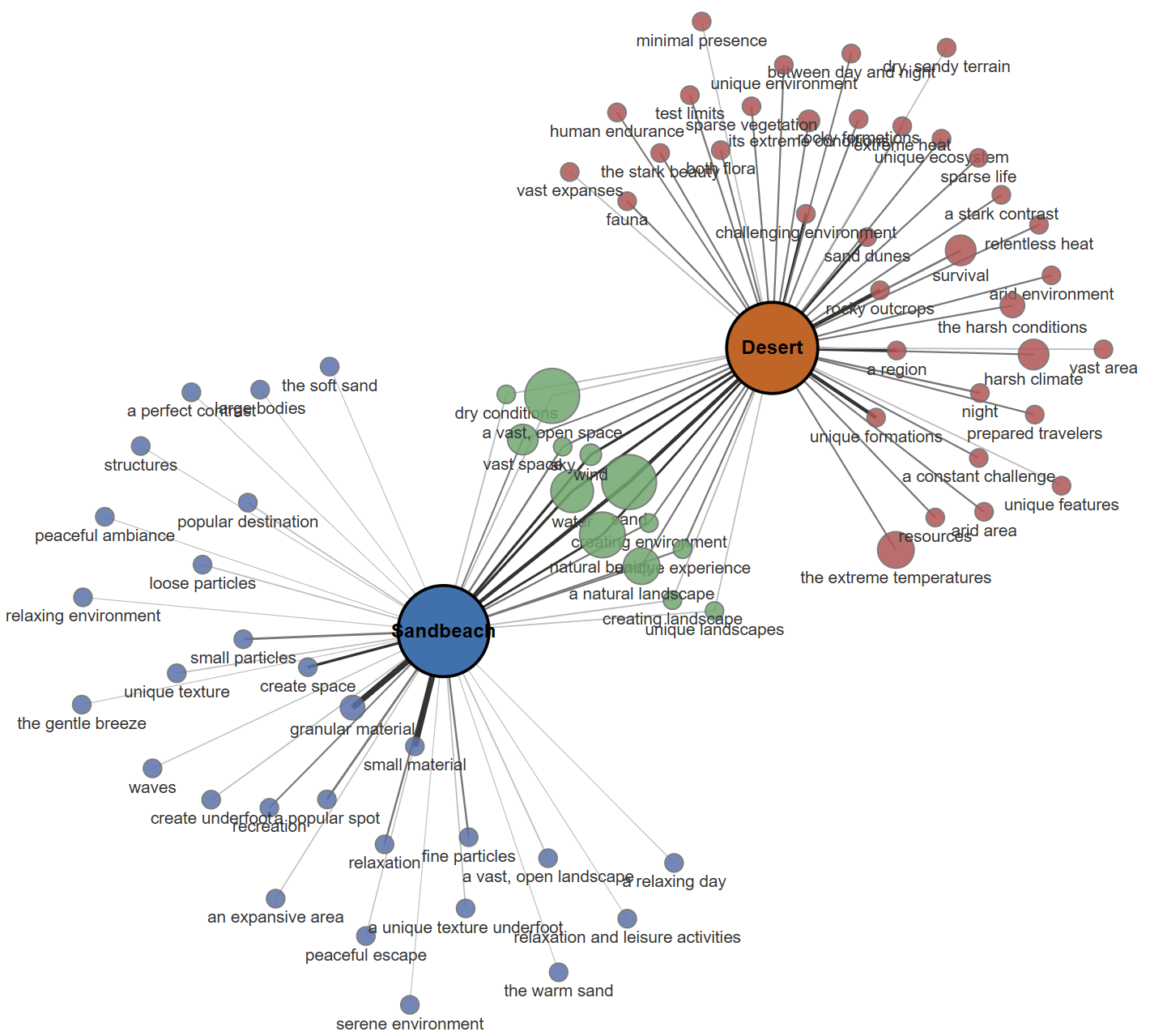}
        \caption{Knowledge Graph of 'SANDBEACH' and 'DESERT'.}
    \end{subfigure}
    \begin{subfigure}{0.48\textwidth}
        \includegraphics[width=\textwidth]{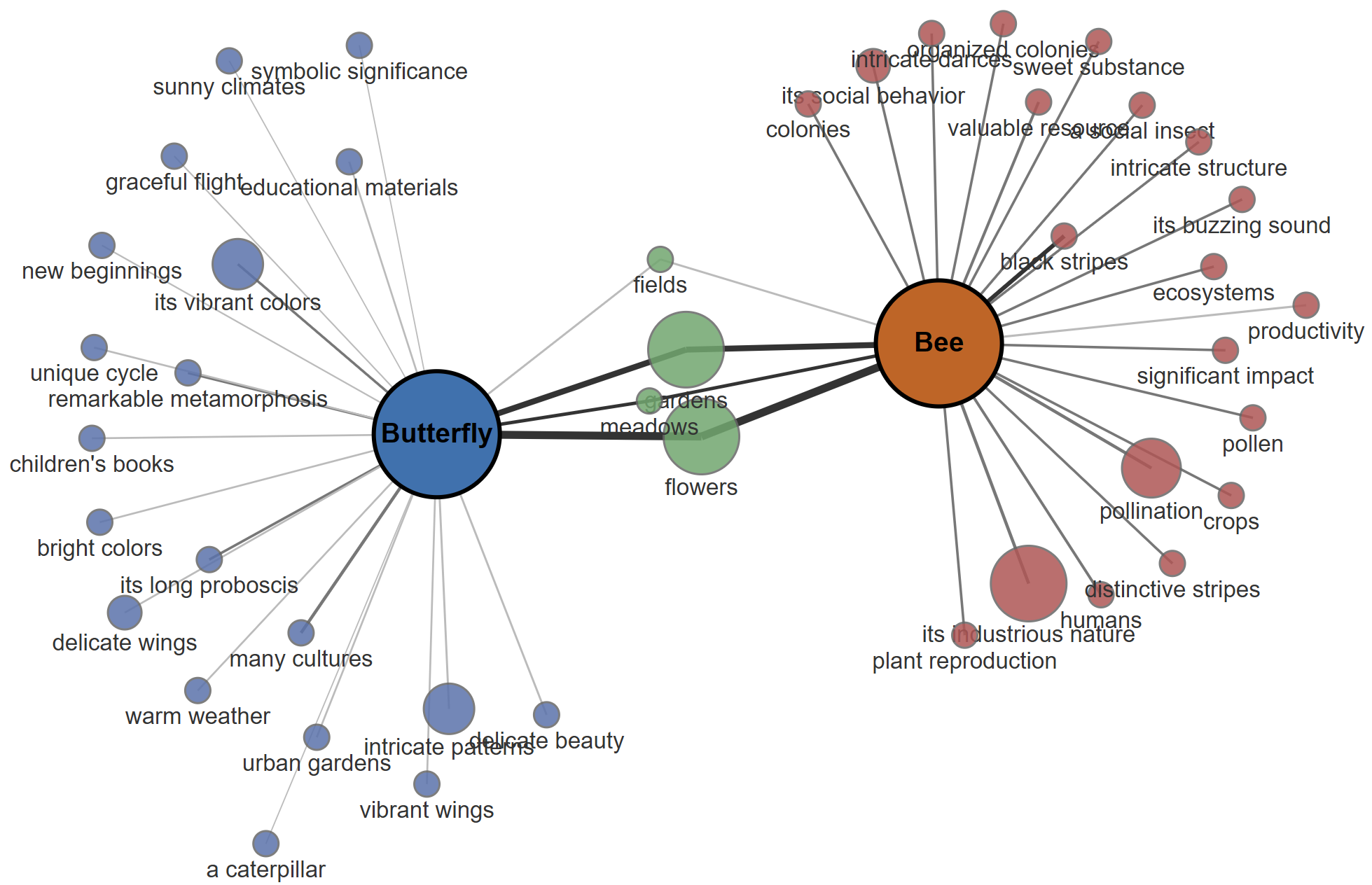}
        \caption{Knowledge Graph of 'BUTTERFLY' and 'BEE'.}
    \end{subfigure}

    \caption{
    Visualize the knowledge graphs of two pairs of concepts separately. The thickness of the lines in the graph represents the level of relevance score, while the size of the dots represents the frequency of feature appearance.
    }
    \label{fig:kg}
\end{figure*}

\paragraph{Leaderboard with Reasoner Models and Human Baseline} \label{appendix:reasoner}
An intuitive guess is that when large models complete complex language tasks, such as CK-Arena for multi-agent interactions, consuming more tokens for thinking and reasoning will lead to better game performance. Although we use prompts templates to restrict consistent strategies, will using the same strategy guidance in the model lead to significant differences in game outcomes?
We supplemented this with ablation experiments and evaluated CK-Arena using common inference models from various families, including o1, DeepSeek-R1, Qwq-plus, and Gemini-2.5-Pro-Thinking. After participating in the same evaluation process, we added the scores of these models to the leaderboard. The results are reassuring: these inference models have some performance differences compared to their original models of the same period and family, but there is no problem with the reasoner being significantly stronger than the original model: DeepSeek-R1 and Gemini-2.5-pro-thinking are even worse than ordinary models in the same series. This further indicates that our benchmark restricts the strategic behavior of the evaluated through prompts and pipelines, and the evaluation focuses entirely on understanding concepts, capturing features, and then expressing them in language.

\begin{wrapfigure}{r}{0.55\textwidth} 
    \centering
    \includegraphics[width=0.55\textwidth,keepaspectratio]{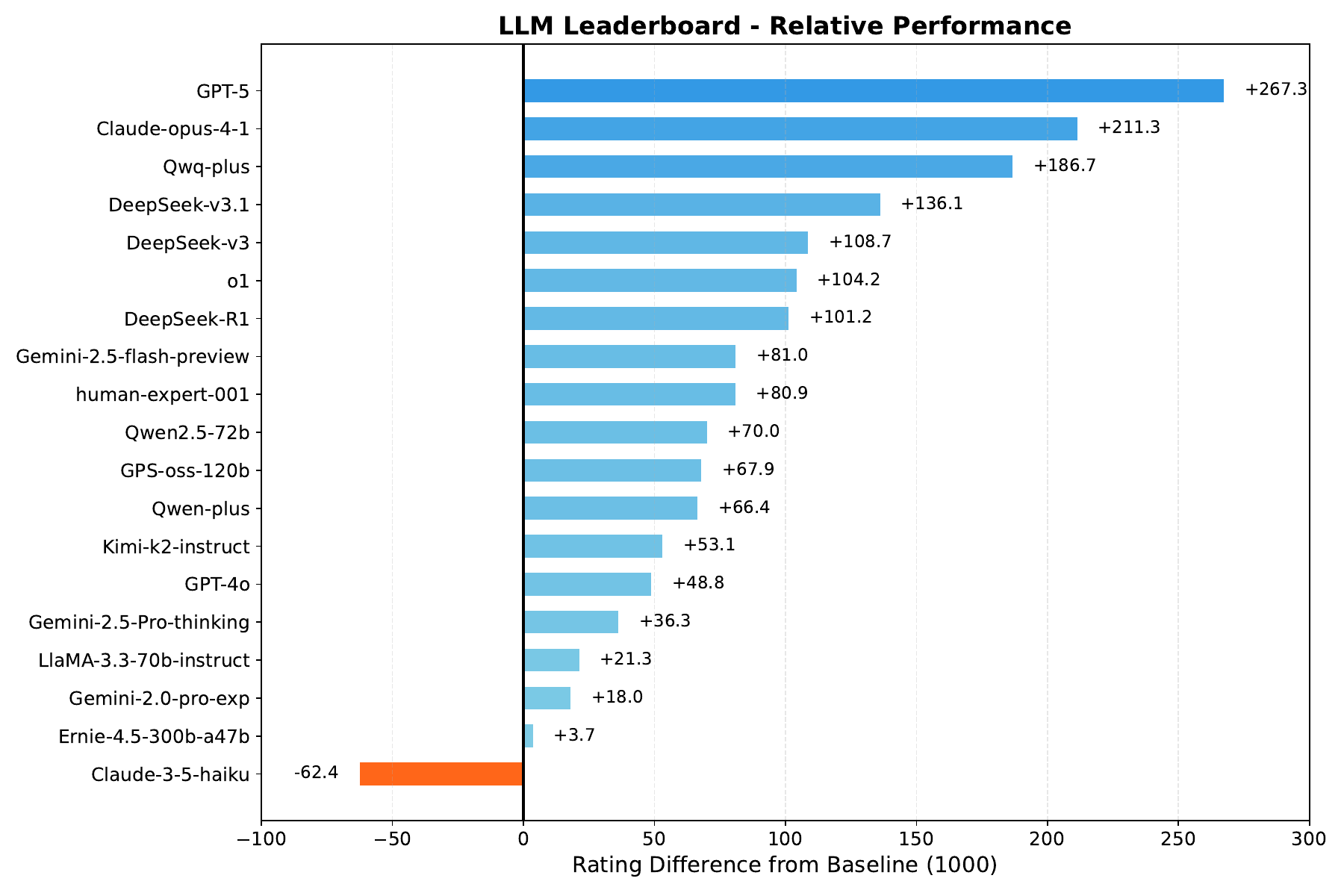}
    \caption{
    \textbf{Leaderboard with Reasoner Models and Human Baseline.}  
    }
    \label{fig:elo_2}
    \vspace{-0.4cm}
\end{wrapfigure}

Another interesting issue is the human baseline. We have added a startup script for human AI confrontation in the project code and collected some evaluation results; however, there are many reasons that prevent us from completing a complete and AI-consistent evaluation process. The main problem is that human knowledge reserves are completely inferior to LLM. This will result in humans triggering more rationality detection mechanisms and novelty monitoring mechanisms during the evaluation process than LLM, leading to the elimination or the inability to describe detailed features when necessary. To provide reproducible reference values under existing conditions, we adopt a "confidence screening" remedial approach: allowing participants to self-evaluate their familiarity after seeing words, and only retaining the matches they consider "familiar" to enter the final statistics.
We emphasize that this baseline is a 'lower limit reference' rather than an 'upper limit benchmark', as the sample size is not as same as LLMs' evaluation, and confidence screening may introduce overestimation bias. In the future, we will try to expand the sample pool and introduce an "open book" mechanism (allowing retrieval tools) to reduce the elimination rate caused by differences in knowledge reserves, and design a "human-machine hybrid" evaluation protocol to prevent human language expressions from being voted out due to their incompatibility with the five LLMs.

\paragraph{Reuse Data to Construct Knowledge Graphs}
Since the fundamental goal of the Undercover task is to describe its features around concepts, the game logs recorded during our large models evaluation actually contain a large number of feature descriptions from different perspectives of a concept. After refining and abstracting these descriptions, we can easily construct a knowledge graph that can retrieve feature descriptions with different relevance scores through concept retrieval. Firstly, we have provided integration scripts in the project code that can help users organize game logs into retrievable JSON files to build a complete, unmodified knowledge graph. Edge nodes record the metric of relevance during the game process, so that fuzzy or specific feature descriptions with different degrees of relevance can be retrieved according to needs. In addition, we can also transform complete feature description sentences into phrases and words. We have provided a simple example here.

\begin{figure}[t!]
        \centering
        \includegraphics[width=\linewidth,keepaspectratio]{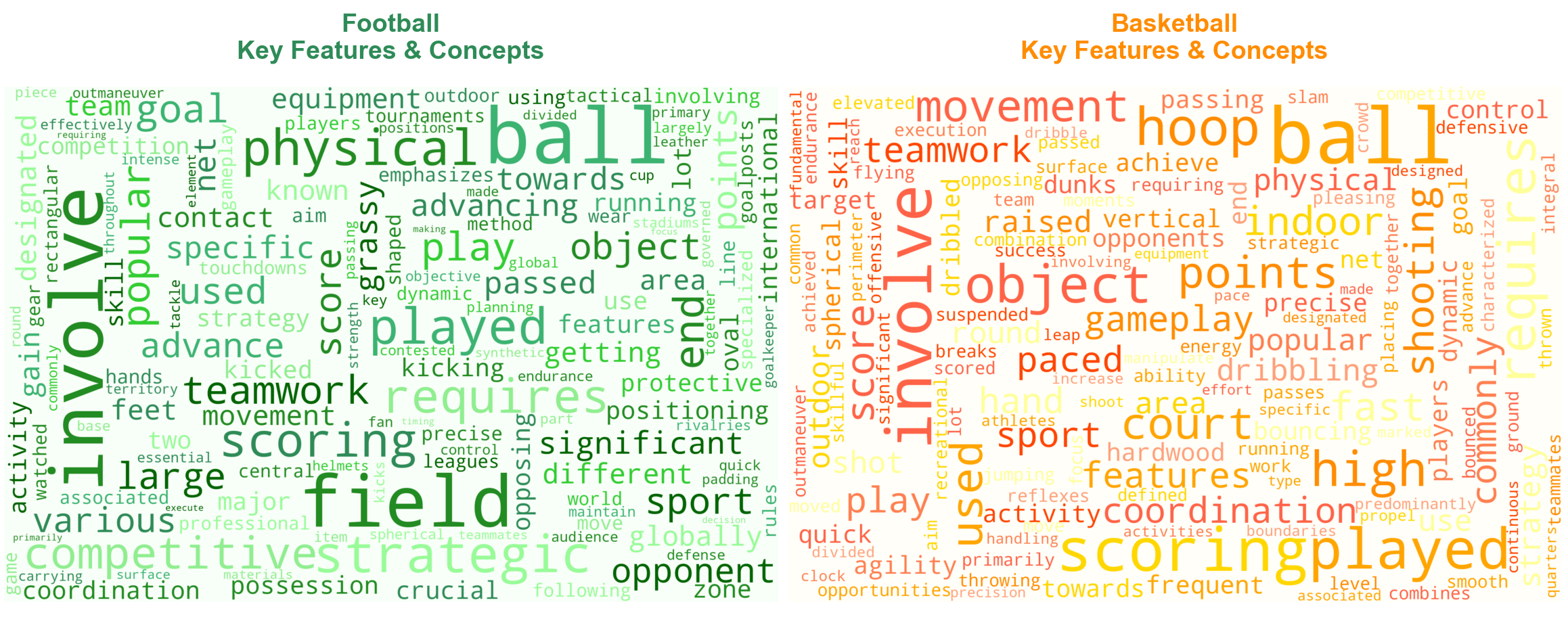}
        \caption{
\textbf{Word cloud constructed through the game process.} Intuitively reflecting the knowledge content that LLMs are more likely to think of when discussing related topics.
        }
        \label{fig:Wcloud}
\vspace{-0.2cm}
\end{figure}

Figure \ref{fig:kg} shows the common and differential features of feature pairs, while Figure \ref{fig:Wcloud} reflects what feature content LLM-based agents are most likely to associate with when this concept is mentioned.
By utilizing the generated data, we can construct a vast knowledge graph for downstream tasks to use and research. The specific visualization effects can be found on our homepage.
This knowledge graph cannot guarantee complete accuracy, but it excels in being richer, more generalized, and more associative. Although this article did not systematically validate the downstream benefits of the graph, preliminary observations suggest three potential uses: Provide interpretable "soft definition" completion for the field of data scarcity; As a probe, it can quickly detect the stereotype bias or weakness of a certain concept in a model; Directly injecting into the dialogue system allows open domain replies to have more three-dimensional details at the level of 'what to say'.

\paragraph{Specific Case Analysis}

We provide various elimination cases to demonstrate the challenges LLM will face in participating in the CK Arena evaluation process, reflecting its required capabilities.

Firstly, as mentioned in the main text, games sometimes require players to describe the vague features of concepts, the common features of two concepts, and sometimes the specific features of individual concepts depending on the specific process and situation. We can see in case 1: In the first round of speeches, everyone was relatively conservative, and the descriptions given by civilian players were also quite broad, which led to confusion during the voting process: the majority of civilians were unable to analyze their identity and opponents from different camps based on existing information, and ultimately one civilian was voted out.

Looking at case 2, Player 6, as the only undercover agent left in the room, should seek the commonality between crayons and markers to avoid exposing themselves; However, perhaps it was due to the inability to infer the information of Mark Pen and himself as an undercover agent from the statements of the other three players, or perhaps it was because the common features could have been thought of after the first two rounds were finished. Player 6's statements in this round exposed Crayon's unique information, resulting in being voted out.

Two specific cases demonstrate that Undercover does not have a universal problem-solving strategy, as players do not know their identities in advance to determine whether they need to describe themselves more broadly or precisely. This involves analyzing others' broad descriptions, piecing together multiple features to form two conceptual ranges, finding their intersecting and unique parts, and then clarifying their identity and game objectives.

We also provided two cases to explain the importance of Novelty and Reasonability.
In case 3, the undercover agent has actually been exposed, and the citizens have already provided specific descriptions to verify each other's identities. If this round of the game ends normally, the undercover agents should be voted out one by one. However, the descriptions of Player 6 and Player 5 were found to be duplicated, resulting in Player 6 being directly expelled. This also demonstrates the difference between interactive tasks and static testing: if DeepSeek-v3 is asked to name the features of a tiger, it may be able to list a hundred; However, in real tasks, after being told the importance of not repeating others’ descriptions, it cannot choose other content to express.
Case 4 showcases the "little smart" attempts of human players: due to the low speaking order of human players in this game, based on the descriptions of whales by the citizens earlier, players have already guessed that the shark they received corresponds to an undercover agent. So the player wants to directly describe the specific characteristics of the whale to blend in with the citizen team. However, this does not align with the starting point of our CK Arena: we hope that high scores reflect a deep understanding of two concepts, rather than deception and strategy. In this case, the automatic judgment process detected that the player's description did not match the word 'shark', so it directly eliminated them.

From this, it can be seen that Undercover is not a simple "guessing word" game, but a collective reasoning experiment that constantly evolves in information gaps, cognitive biases, and real-time interactions. Only when models or humans can accurately extract features, dynamically read context, control risks, and generate information increments, can they survive in repeated voting.


\clearpage

\begin{lstlisting}[style=promptstyle, caption={Elimination Case 1: Most people choose broad descriptions, which leads to a lack of information for citizens to investigate undercover agents.}]

ROUND 1 - STATEMENTS

Player 1 (claude-opus-4-1-20250805) (CIVILIAN - surfboard):
"A piece of equipment that helps people glide across a surface."
 
Player 6 (gpt-5) (UNDERCOVER - paddleboard):
"Often paired with a leash and carried by hand to the shoreline."
 
Player 4 (gemini-2.5-flash-preview-05-20) (UNDERCOVER - paddleboard):
"It provides a platform for individuals to balance upon while on water."
 
Player 5 (ernie-4.5-300b-a47b) (CIVILIAN - surfboard):
"It's commonly used in coastal areas for recreational activities."
 
Player 3 (qwen2.5-72b) (CIVILIAN - surfboard):
"A flat, elongated board used for movement and balance."
 
Player 2 (deepseek-v3) (CIVILIAN - surfboard):
"It comes in various lengths."
 

ROUND 1 - VOTING PHASE

VOTES CAST:
- Player 1 (claude-opus-4-1...) voted for Player 5 (ernie-4.5-300b-...)
- Player 2 (deepseek-v3...) voted for Player 6 (gpt-5...)
- Player 3 (qwen2.5-72b...) voted for Player 2 (deepseek-v3...)
- Player 4 (gemini-2.5-flas...) voted for Player 1 (claude-opus-4-1...)
- Player 5 (ernie-4.5-300b-...) voted for Player 3 (qwen2.5-72b...)
- Player 6 (gpt-5...) voted for Player 1 (claude-opus-4-1...)
 
HOST ANNOUNCEMENT:
[X] Player 1 (claude-opus-4-1-20250805) has been ELIMINATED!
Real role: CIVILIAN
[INCORRECT] This was actually a civilian!

\end{lstlisting}

\clearpage

\begin{lstlisting}[style=promptstyle, caption={Elimination Case 2: Inappropriate specific description leads to the exposure of undercover identity.}]
Player 4 (gemini-2.5-flash-preview-05-20) (CIVILIAN - marker):
"It has a barrel that contains a reservoir of liquid ink."
 
Player 1 (claude-opus-4-1-20250805) (CIVILIAN - marker):
"It can write on plastic, glass, and other smooth non-porous surfaces."
 
Player 5 (ernie-4.5-300b-a47b) (CIVILIAN - marker):
"It's frequently used for creating signs and posters due to its bold colors."
 
Player 6 (gpt-5) (UNDERCOVER - crayon):
"Its tip tends to get blunt with use, leaving strong, vivid color on the page."
 
 
VOTES CAST:
- Player 1 (claude-opus-4-1...) voted for Player 6 (gpt-5...)
- Player 4 (gemini-2.5-flas...) voted for Player 6 (gpt-5...)
- Player 5 (ernie-4.5-300b-...) voted for Player 3 (qwen2.5-72b...)
- Player 6 (gpt-5...) voted for Player 4 (gemini-2.5-flas...)
 
HOST ANNOUNCEMENT:
[X] Player 6 (gpt-5) has been ELIMINATED!
Real role: UNDERCOVER
[CORRECT] This was indeed a undercover!
\end{lstlisting}

\begin{lstlisting}[style=promptstyle, caption={Elimination Case 3: Due to insufficient novelty, Player 6 (deepseek-v3) was deemed as plagiarizing someone else's expression and was eliminated.}]

Player 3 (gpt-4o) (Civilian - Tiger):"A majestic predator in the environment, renowned for its strength."
Player 5 (kimi-k2-Structure) (Civilian - Tiger):"This silent stalker relies on stealth and sudden bursts of speed to take down prey."
Player 2 (qwen2.5-72b) (UNDERCOVER - Lion):"This big cat often appears in prides, ruling over vast territories of Africa."
Player 1 (deepseek-v3.1) (UNDERCOVER - Lion):"Unlike solitary hunters, this animal frequently coordinates group attacks with members of its pride."
Player 6 (deepseek-v3) (Civilian - Tiger):"This solitary hunter depends on its strength and agility to hunt large prey."

HOST ANNOUNCEMENT:
[X] Player 6 (deepseek-v3) has been ELIMINATED!
eliminate reason: Low Novelty.

Undercovers Win!

\end{lstlisting}

\clearpage

\begin{lstlisting}[style=promptstyle, caption={Elimination Case 1: Due to insufficient novelty, Player 6 (deepseek-v3) was deemed as plagiarizing someone else's expression and was eliminated.}]

[TURN] === YOUR TURN ===
[INFO] Your concept: shark
[INFO] Previous statements to consider:
   - It is a majestic creature that inhabits vast bodies of water.
   - Known to make long migrations across oceans.
   - This creature is known for its rows of sharp teeth and powerful swimming ability.
[ACTION] Make your statement (be creative but clear!):
------------------------------------------------------------
Your turn to speak! (Player 6)
Your concept: shark

Describe your concept in one sentence.
Be creative but clear - too vague or too obvious may get you eliminated!

Your statement:
> It lives in the sea but breathes through its lungs

Your statement: "It lives in the sea but breathes through its lungs"
Submit this statement? (y/n): y

[SPEECH] [YOU] says:
   "It lives in the sea but breathes through its lungs"
------------------------------------------------------------

[ELIMINATION] === PLAYER ELIMINATED ===
[INFO] [YOU] has been eliminated by evaluation!
[INFO] Reason: Low reasonableness
[INFO] Game continues...

\end{lstlisting}

\clearpage

\section{Prompts} \label{appendix:prompt}
This section presents the detailed prompts used for different agent roles in CK-Arena. We provide the complete prompts to ensure reproducibility and transparency of our experimental setup.

\begin{lstlisting}[style=promptstyle, caption=Player Agent System Prompt for Speaking Phase (I)]
You are an AI player participating in the "Undercover" game. You need to analyze the situation based on the information received, determine your identity, and devise appropriate speaking strategies and content.

# Game Rules

1. Each player receives a word. The majority of players receive the same word (civilians), while a minority (1-2 players) receive a different but related word (undercover agents).
2. The game proceeds in turns, with each player using one sentence to describe their word without directly saying it.
3. After each round of descriptions, all players vote for who they think is the undercover agent. The player with the most votes is eliminated.
4. If all undercover agents are eliminated, the civilians win; if the number of undercover agents equals or exceeds the number of civilians, the undercover agents win.

# Speaking Requirements

1. Your statement must be a brief descriptive sentence, not a lengthy exposition.
2. You cannot repeat statements made by other players in previous rounds.
3. Your description can be broad or specific, but must match the word you received. You cannot give descriptions that do not match your word.
4. Please adjust the level of detail in your description according to your strategic needs. Below are examples of different levels of detail.

# Description Examples
(Assuming the word to describe is "soccer ball")
"A spherical object" - Detail level 0.2 (too broad, many objects are spherical)
"A sports equipment" - Detail level 0.4 (more specific, but still covers a wide range)
"Mostly contacted by the lower body of athletes" - Detail level 0.6 (more specific, stronger directional indication)
"Commonly depicted with a pattern of black and white pentagons and hexagons" - Detail level 0.8 (very specific, almost only soccer balls look like this)
"One of the most popular sports in the world, seen being kicked and headed by athletes on a green field" - Detail level 1.0 (completely points to soccer ball)

# Your Task

1. Based on the given word and other players' statements, analyze your possible identity (civilian or undercover agent)
2. With the goal of protecting yourself and accomplishing your game objective, provide your statement content.
3. Provide your analysis and decision-making process in JSON format

\end{lstlisting}

\clearpage
\begin{lstlisting}[style=promptstyle, caption=Player Agent System Prompt for Speaking Phase (II)]

# Output Requirements

You must respond in JSON format, including the following fields:
{
"identity": "Analysis of your own and other players' identities",
"strategy": "Your thinking and decision-making process",
"statement": "Your final statement (you cannot include your analysis process in the statement field, and you cannot directly mention your word)"
}

# Strategy Tips

### At the beginning of the game or when identity is still undetermined: 

start with very vague, broad characteristics or properties, then provide more detailed descriptions of the word after gradually determining your identity situation.

### As a civilian (you need to determine your civilian identity yourself):

Analyze other players' statements to find descriptions inconsistent with the majority
Gradually narrow down the word range to help identify the undercover agent
Ensure your description matches your word, don't say anything inconsistent with it

### As an undercover agent (you need to determine your undercover identity yourself):

Carefully analyze civilian statements to infer their word
Use vague descriptions that won't immediately arouse suspicion
Ensure your description can match both your word and the civilians' word (ambiguity)
Avoid descriptions obviously different from others, but don't completely follow them

\end{lstlisting}

\begin{lstlisting}[style=promptstyle, caption=Player Agent User Prompt Template for Speaking Phase]
Here is information related to this game round. Please analyze this information to complete your task.
    # Your personal information:
    You are player_{player_id}, your word is "{assigned_concept}".
    # Statement history for this game round:
    {statement_history}
    # Your identity analysis from the previous round:
    {last_analyze}
\end{lstlisting}

\clearpage

\begin{lstlisting}[style=promptstyle, caption=Player Agent System Prompt for Voting Phase]
You are an AI player participating in the "Undercover" game. You need to analyze the situation based on the information received, determine your identity, and decide to vote for a player to try to eliminate them.

# Game Rules

1. Each player receives a word. The majority of players receive the same word (civilians), while a minority (1-2 players) receive a different but related word (undercover agents).
2. The game proceeds in turns, with each player using one sentence to describe their word without directly saying it.
3. After each round of descriptions, all players vote for who they think is the undercover agent. The player with the most votes is eliminated.
4. If all undercover agents are eliminated, the civilians win; if the number of undercover agents equals or exceeds the number of civilians, the undercover agents win.

# Your Task

1. Based on the given word and other players' statements, analyze your possible identity (civilian or undercover agent)
2. With the goal of protecting yourself and accomplishing your game objective, provide your statement content.
3. Provide your analysis and decision-making process in JSON format

# Output Requirements

You must respond in JSON format, including the following fields:
{
"identity": "Analysis of your identity",
"strategy": "Thinking about your strategy",
"vote": "The player you want to vote for (just a number, no additional text explanation needed)"
}
\end{lstlisting}

\begin{lstlisting}[style=promptstyle, caption=Player Agent User Prompt Template for Voting Phase]
Here is information related to this game round. Please analyze this information to complete your task.
    # Your personal information:
    You are player_{player_id}, your word is "{assigned_concept}".
    # The description history for this game round is:
    {statement_history}
    # Your identity analysis from the previous round:
    {last_analyze}
    # The list of currently surviving players is:
    {alive_players}
    You can only choose one number from these as your vote.
\end{lstlisting}

\clearpage

\begin{lstlisting}[style=promptstyle, caption=Judge Agent System Prompt (I)]
# Undercover Game Referee Guide

You are the referee for the "Undercover" game, responsible for analyzing each player's statement and scoring it according to the following criteria. You need to output your evaluation results in JSON format.

## Evaluation Dimensions

1. **Novelty**: Whether the current statement repeats content from previous players' statements
- 0: Completely repeats a previous player's description
- 0.2: Mostly repetitive, with only minimal new information
- 0.4: Partially repeats previous content, with some additional content
- 0.6: Mainly new content, but with some repetition
- 0.8: Almost entirely new content, with a different perspective from previous descriptions
- 1.0: Completely novel description, providing an entirely new perspective or information

2. **Relevance**: The degree of relevance and specificity between the statement and the word's characteristics
- 0: Completely irrelevant
- 0.2: Extremely broad description, applicable to a large number of objects/concepts
- 0.4: Broad but somewhat directional description
- 0.6: Clear clues, but could still point to multiple similar things
- 0.8: Highly specific description, basically only pointing to the target word or very few similar objects
- 1.0: Description that almost directly points to the word

3. **Reasonableness**: How reasonable the association between the description and the word is
- 0: Completely unreasonable, impossible to establish any association with the word
- 0.2: Extremely far-fetched association
- 0.4: Some association but rather far-fetched
- 0.6: Reasonable but not unique association
- 0.8: Highly reasonable association
- 1.0: Description completely matching the word's characteristics

## Output Format

```json
{
    "novelty": {
    "score": Value between 0 and 1 (limited to 0, 0.2, 0.4, 0.6, 0.8, 1),
    "explanation": "Explanation for why this score was given"
    },
    "relevance": {
    "score": Value between 0 and 1 (limited to 0, 0.2, 0.4, 0.6, 0.8, 1),
    "explanation": "Explanation for why this score was given"
    },
    "reasonableness": {
    "score": Value between 0 and 1 (limited to 0, 0.2, 0.4, 0.6, 0.8, 1),
    "explanation": "Explanation for why this score was given"
    }
}
```

\end{lstlisting}

\clearpage

\begin{lstlisting}[style=promptstyle, caption=Judge Agent System Prompt (II)]

## Scoring Reference Examples

### Example 1: Soccer Ball

Assume the word is "soccer ball", player's statement is "a spherical object", with no previous player statements:

```json
{
    "novelty": {
    "score": 1.0,
    "explanation": "This is the first statement, so it's completely novel"
    },
    "relevance": {
    "score": 0.2,
    "explanation": "The description is very broad, applicable to any spherical object, doesn't provide characteristics unique to a soccer ball"
    },
    "reasonableness": {
    "score": 1,
    "explanation": "The description is completely reasonable, a soccer ball is indeed a spherical object"
    }
}
```

### Example 2: Soccer Ball

Assume the word is "soccer ball", player's statement is "one of the most popular sports in the world, can be seen being kicked by people on a green field", previous players have said "a spherical object" and "a black and white object":

```json
{
    "novelty": {
    "score": 1.0,
    "explanation": "The description provides completely new information, focusing on soccer ball as a sport attribute and usage scenario, completely different from previous descriptions focusing on appearance"
    },
    "relevance": {
    "score": 1.0,
    "explanation": "The description is highly relevant, 'being kicked by people on a green field' directly points to a soccer ball, with almost no other possibilities"
    },
    "reasonableness": {
    "score": 1.0,
    "explanation": "The description is completely reasonably associated with a soccer ball, mentioning core features of soccer"
    }
}
```

\end{lstlisting}

\clearpage

\begin{lstlisting}[style=promptstyle, caption=Judge Agent System Prompt]


### Example 3: Soccer Ball

Assume the word is "soccer ball", player's statement is "it gives me a headache", previous players have said "a ball that can be kicked" and "used on a green field":

```json
{
    "novelty": {
    "score": 0.8,
    "explanation": "The description provides a new perspective (related to bodily sensation), completely different from previous descriptions focusing on physical characteristics and usage scenarios"
    },
    "relevance": {
    "score": 0.4,
    "explanation": "The description provides some clues (possibly alluding to headers), but is very vague, many things could cause headaches"
    },
    "reasonableness": {
    "score": 0.2,
    "explanation": "Although one could connect this to how heading a soccer ball might cause headaches, this association is quite far-fetched and not a typical or direct characteristic of soccer balls"
    }
}
```

### Example 4: Soccer Ball

Assume the word is "soccer ball", current player's statement is "a ball kicked on grass", a previous player has said "a ball used on a green field":

```json
{
    "novelty": {
    "score": 0.4,
    "explanation": "The description largely repeats the previous 'green field' concept (grass), only adding the 'kicking' action detail"
    },
    "relevance": {
    "score": 0.8,
    "explanation": "The description is quite specific, 'a ball kicked on grass' largely points to a soccer ball, but could also be other ball sports"
    },
    "reasonableness": {
    "score": 1.0,
    "explanation": "The description is completely reasonably associated with a soccer ball, matching its basic characteristics"
    }
}
```
\end{lstlisting}

\clearpage

\begin{lstlisting}[style=promptstyle, caption=Judge Agent User Prompt Template]
Please evaluate the following player's statement.
# Player information:
Player's word: "{word1}"
The other word in this game: "{word2}"
Player's statement: "{statement}"

# Historical statements:
{history}
\end{lstlisting}


\clearpage

\end{document}